\documentclass[sigconf, nonacm]{acmart}

\newcommand\vldbdoi{XX.XX/XXX.XX}

\newcommand\vldbavailabilityurl{\repoURL}
\newcommand\vldbpagestyle{plain} 

\usepackage{amsmath,amsfonts}
\usepackage{graphicx}
\usepackage{textcomp}
\usepackage{xcolor}

\usepackage[ruled,linesnumbered,noend]{algorithm2e}
\usepackage{etoolbox}
\makeatletter
\patchcmd{\@algocf@start}
  {-1.5em}
  {0pt}
  {}{}
\makeatother

\makeatletter
\def\blfootnote{\gdef\@thefnmark{}\@footnotetext}
\makeatother

\usepackage{balance}
\usepackage{url}
\usepackage{multirow}
\usepackage{multicol}
\usepackage{listings}
\usepackage{caption}
\usepackage{wrapfig}
\usepackage{soul}
\PassOptionsToPackage{table,xcdraw}{xcolor}
\usepackage{fancyvrb}
\usepackage{eepic}
\usepackage[version=0.96]{pgf}
\usepackage[latin1]{inputenc}
\usepackage{verbatim}
\usepackage{moreverb}
\usepackage{epsfig}
\usepackage{subfig}
\usepackage{bm}
\usepackage[section]{placeins}
\usepackage{enumitem}
\usepackage{booktabs} 
\usepackage{makecell}
\usepackage[export]{adjustbox}
\usepackage{fancyvrb}
\usepackage{varwidth}

\usepackage[english]{babel}
\def\ncp{\vspace*{-0.5ex}}

\def\shorten{\looseness=-1} 
\newcommand{\nsstitle}[1]{\noindent\textup{\textbf{#1}}}

\newcommand{\sysname}{KGLiDS} 
\newcommand{\graphname}{LiDS}
\newcommand{\repoURL}{https://github.com/CoDS-GCS/kglids}
\newcommand{\holoclean}{HoloClean}

\newcommand{\autolearn}{AutoLearn}

\newcommand{\totalDatasets}{51}

\newcommand{\myNum}[1]{(\emph{#1})}

\DeclareMathOperator*{\concat}{%
    \mathchoice%
        {\Big\Vert}%
        {\big\Vert}%
        {\Vert}%
        {\Vert}%
}

\begin{document}

\title{KGLiDS: A Platform for Semantic Abstraction, Linking, and Automation of Data Science \\ \   [Technical Report]}

\author{Mossad Helali$^1$, Niki Monjazeb$^1$, Shubham Vashisth$^1$, Philippe Carrier$^1$, Ahmed Helal$^1$, \\ Antonio Cavalcante$^2$, Khaled Ammar$^3$, Katja Hose$^4$, Essam Mansour$^1$}
\affiliation{%
  \institution{$^1$\textit{Concordia University} \qquad\qquad $^2$\textit{Borealis AI}   \qquad\qquad $^3$\textit{University of Waterloo}\qquad\qquad  $^4$\textit{TU Wien} 
  \\
  \qquad Canada \qquad\qquad\qquad\qquad Canada   \qquad\qquad\quad\qquad\qquad Canada   \qquad\quad\qquad\qquad Austria
}
}



\begin{abstract}
In recent years, we have witnessed the growing interest from academia and industry in applying data science technologies to analyze large amounts of data. In this process, a myriad of artifacts (datasets, pipeline scripts, etc.) are created. However, there has been no systematic attempt to holistically collect and exploit all the knowledge and experiences that are implicitly contained in those artifacts. Instead, data scientists recover information and expertise from colleagues or learn via trial and error. Hence, this paper presents a scalable platform, KGLiDS, that employs machine learning and knowledge graph technologies to abstract and capture the semantics of data science artifacts and their connections. Based on this information, KGLiDS enables various downstream applications, such as data discovery and pipeline automation. Our comprehensive evaluation covers use cases in data discovery, data cleaning, transformation, and AutoML. It shows that KGLiDS is significantly faster with a lower memory footprint than the state-of-the-art systems while achieving comparable or better accuracy.\shorten  
\end{abstract}

\maketitle

\blfootnote{ $^\ast$For emails please contact: mossad.helali@mail.concordia.ca}


\pagestyle{\vldbpagestyle}
\begingroup\small\noindent
\endgroup
\begingroup
\endgroup



\section{Introduction}
\label{sec:introduction}
Data science is the process of collecting, cleaning, and analyzing structured and unstructured data to derive insights and make predictions. To define appropriate pipelines, data scientists dedicate significant time to data discovery, data preparation, and modeling. Unfortunately, data scientists primarily work in isolation without exchanging knowledge with fellow data scientists working on similar tasks and with similar datasets. Consequently, there is only little to no exchange of knowledge and experiences, a phenomenon referred to as ``tribal knowledge''.  
Interestingly, scientists can access valuable pipelines not only within individual companies but also through open machine learning (ML) portals, such as Kaggle~\cite{kaggle} and OpenML~\cite{OpenML}. These platforms offer entry to extensive public repositories containing thousands of datasets and hundreds of thousands of pipelines with ample opportunities for learning and benefit.\shorten

If it was possible to learn efficiently from such pipelines, data scientists would no longer be forced to reinvent the wheel and instead be able to focus on solving new and complex problems. A crucial prerequisite for realizing this objective of effectively sharing and leveraging the implicit knowledge and experiences embedded in existing data science pipelines~\cite{KEK} is to move beyond the specifics of datasets and pipelines. The key is to abstract from these details and, instead, capture their semantics. This approach facilitates overcoming challenges related to differences in data formats (CSJ, JSON, etc.), programming languages, and libraries. In the context of this paper, we employ knowledge graphs (KGs) to efficiently capture such semantics, relationships, and dependencies -- both with respect to \emph{datasets} and \emph{pipelines}.

\begin{figure*}[t]
 \centering
  \includegraphics[width=\textwidth]{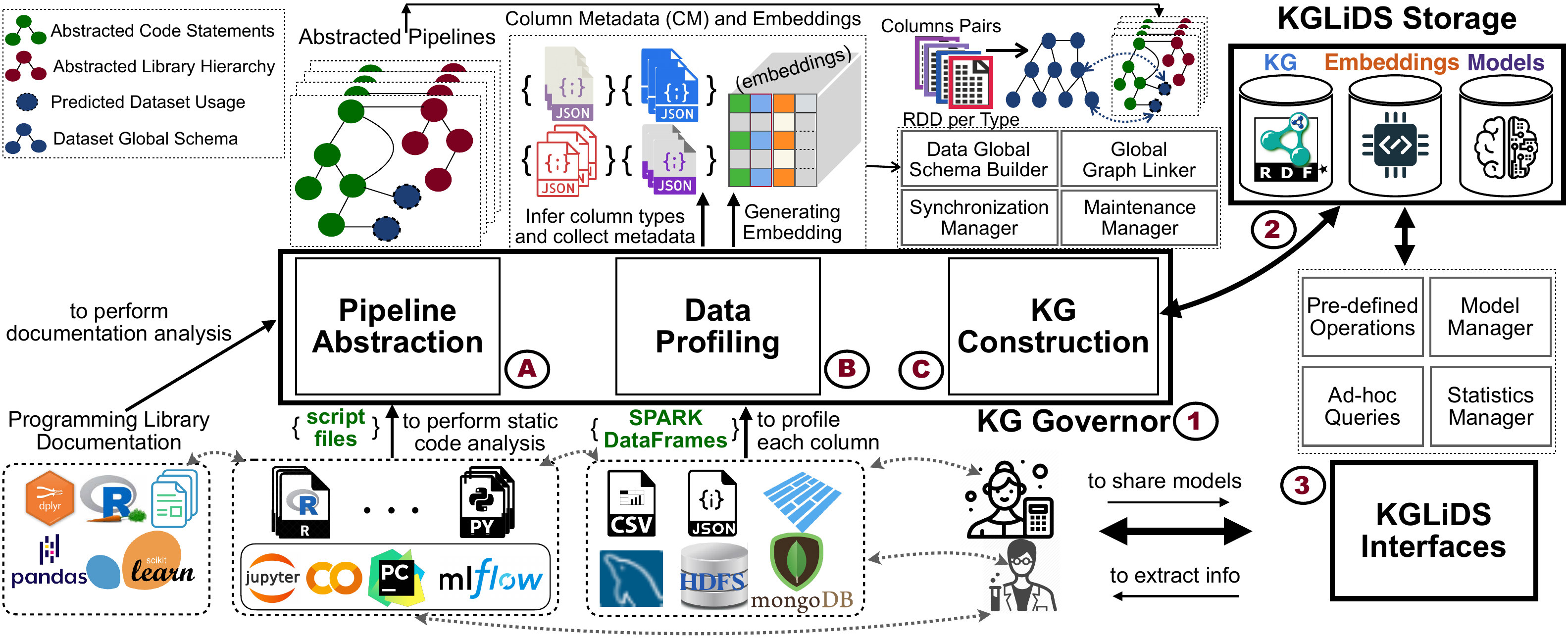} 
  \caption{
  An overview of {\sysname}'s main components: 1) KG Governor for pipeline abstraction, data profiling and KG construction as discussed in Section~\ref{sec:governor}, 2) {\sysname} Storage for our RDF-star KG, embeddings and GNN models, 3) {\sysname} Interfaces, a Python library for different use cases, such as data discovery, cleaning, and transformation, as discussed in Section~\ref{sec:applications}. {\sysname} supports predefined APIs and ad-hoc queries via SPARQL queries. Our library enables users to review recommended operations and execute them. {\sysname} enables automatic learning and discovery on open data science.\shorten}
  \label{fig:overview}
\end{figure*}

Overall, we identify two streams of related work (dataset search and pipeline generation/recommendation), which have so far been considered separately. To the best of our knowledge, KGLiDS is the first approach to combine these two perspectives within a single holistic framework.
Hence, existing data discovery systems~\cite{starmie, santos, d3l, aurum, Fatemeh18, ZhuDNM19} and data science platforms do not capture the semantics of data science artifacts or interlink the semantics of datasets in data lakes with their corresponding semantics of pipeline scripts available in code repositories~\cite{KEK}. 
While there are several systems recommending data preparation operations, such as exploratory data analysis (EDA)~\cite{dataprep}, data transformations~\cite{autolearn, lfe}, or data cleaning~\cite{aimnet, HoloClean, DataWig, kindi2021}, these systems perform excessive analysis on the raw datasets. Hence, they are computationally expensive and suffer limited scalability with large datasets. 

Furthermore, Auto-Suggest~\cite{Auto-Suggest} learns from a collection of data science notebooks to recommend operations, such as Join, Pivot, Unpivot, and GroupBy -- but does not offer support for more complex tasks. AutoML systems, such as~\cite{autosklearn, flaml}, learn to predict mainly a classifier and perform hyper-parameter tuning. 
Nevertheless, these systems are limited to analyzing raw datasets and code and, therefore, cannot capture and exploit the semantics of data science artifacts. 
Although there are a few techniques~\cite{graph4code, code_ontology, hmm_java_ontology} to provide machine-readable semantic abstractions of software code, the majority of these systems target statically-typed languages such as Java, the vast majority of data science pipelines are written in Python, a dynamic language, for which accurate static analysis is challenging or even infeasible in some cases~\cite{pycg}.

This paper addresses these challenges and proposes an approach tailored specifically to the requirements of data science artifacts. We developed  {\sysname}, a fully-fledged platform powered by knowledge graphs capturing the semantics of data science artifacts, including both datasets and pipelines, as well as their interconnections. To the best of our knowledge, {\sysname} is the first platform to combine both dataset and pipeline discovery in a holistic approach. 

We identified popular use cases related to data discovery, data cleaning for null values, data transformation for normalizing column values, and AutoML for predicting an ML classifier or regressor. Moreover, we identified the state-of-the-art (SOTA) systems for these use cases and conducted a comprehensive evaluation against them. Our system demonstrates competitive accuracy with more efficient resource utilization.
In summary, the contributions of this paper are:\shorten
\begin{itemize}
 \item The first and fully-fledged platform\footnote{The {\sysname} repository, ontology, and datasets can be accessed at \url{https://github.com/CoDS-GCS/kglids}} to automate learning and discovery based on the semantics of data science artifacts (the architecture is described in Section~\ref{sec:overview}).

 \item Scalable embedding-based data profiling techniques and fine-grained type inference methods to construct a global representation of datasets (Section~\ref{sec:governor}).
 
 \item Capturing the semantics of data science pipelines at scale with specific support for Python (Section~\ref{sec:governor}).

 \item A novel formalization and scalable implementation of data cleaning and transformation as graph neural network (GNN) classification tasks based on the semantics of data science artifacts and dataset embeddings (Section~\ref{sec:applications}).\shorten

 \item The {\sysname} interfaces with comprehensive pre-defined operations and APIs to access the KG, embeddings, and models. Most of the APIs are implemented using SPARQL (Section~\ref{sec:interfaces}).\shorten

 \item A comprehensive evaluation using datasets used in existing data discovery benchmarks, data cleaning, transformation, and code abstraction. Our experiments show the superiority of {\sysname} over the SOTA systems in terms of time and memory while achieving comparable or better accuracy (Section~\ref{sec:evaluation}).\shorten
\end{itemize}


\section{The {\sysname} Architecture}
\label{sec:overview}

The architecture of {\sysname} is illustrated in Figure~\ref{fig:overview}. The main components are: (1) \emph{KG Governor}, which is responsible for creating, maintaining, and synchronizing the {\sysname} knowledge graph with the covered data science artifacts, pipelines, and datasets, (2) \emph{{\sysname} Storage}, which stores the constructed knowledge graphs as well as embeddings and ML models, (3) \emph{{\sysname} Interfaces}, which allows a diverse range of users to interact with {\sysname} to extract information or share their findings with the community. 

\subsection{The KG Governor and Data Science Artifacts} 
{\sysname} captures the semantics of data science artifacts, i.e., pipelines and their associated datasets, by applying novel methods for pipeline abstraction and data profiling. 
During bootstrapping, KGLiDS is deployed by enabling the KG Governor to profile the local datasets and abstract pipeline scripts to construct a knowledge graph, as illustrated in Figure \ref{fig:overview}.
The \emph{KG Governor} consists of: (A) \emph{Pipeline Abstraction}, which captures the semantics of pipelines by analyzing pipeline scripts, programming libraries documentation, and usage of datasets, (B) \emph{Data Profiling}, which collects metadata and learns representations of datasets including columns and tables, and (C) \emph{Knowledge Graph Construction}, which builds and maintains our KG and embeddings. In \emph{KG Construction}, the \emph{Global Graph Linker} performs link prediction between nodes, e.g., linking a table used in a pipeline and exists in a dataset. 
 
{\sysname} adopts embedding-based methods to predict relationships among data items and utilizes Spark-based approach to guarantee a scalable approach for constructing the graph from large growing datasets and pipelines. 
{\sysname} is not a static platform; as more datasets and pipelines are added, {\sysname} continuously and incrementally maintains our KG. 
To avoid having to run the prediction every time a user wants to use the system, predictions are materialized and the resulting nodes and edges annotated 
with a prediction score expressing the degree of confidence in the prediction.\shorten 

\begin{figure}[t]
 \centering
  \includegraphics[width=\columnwidth]{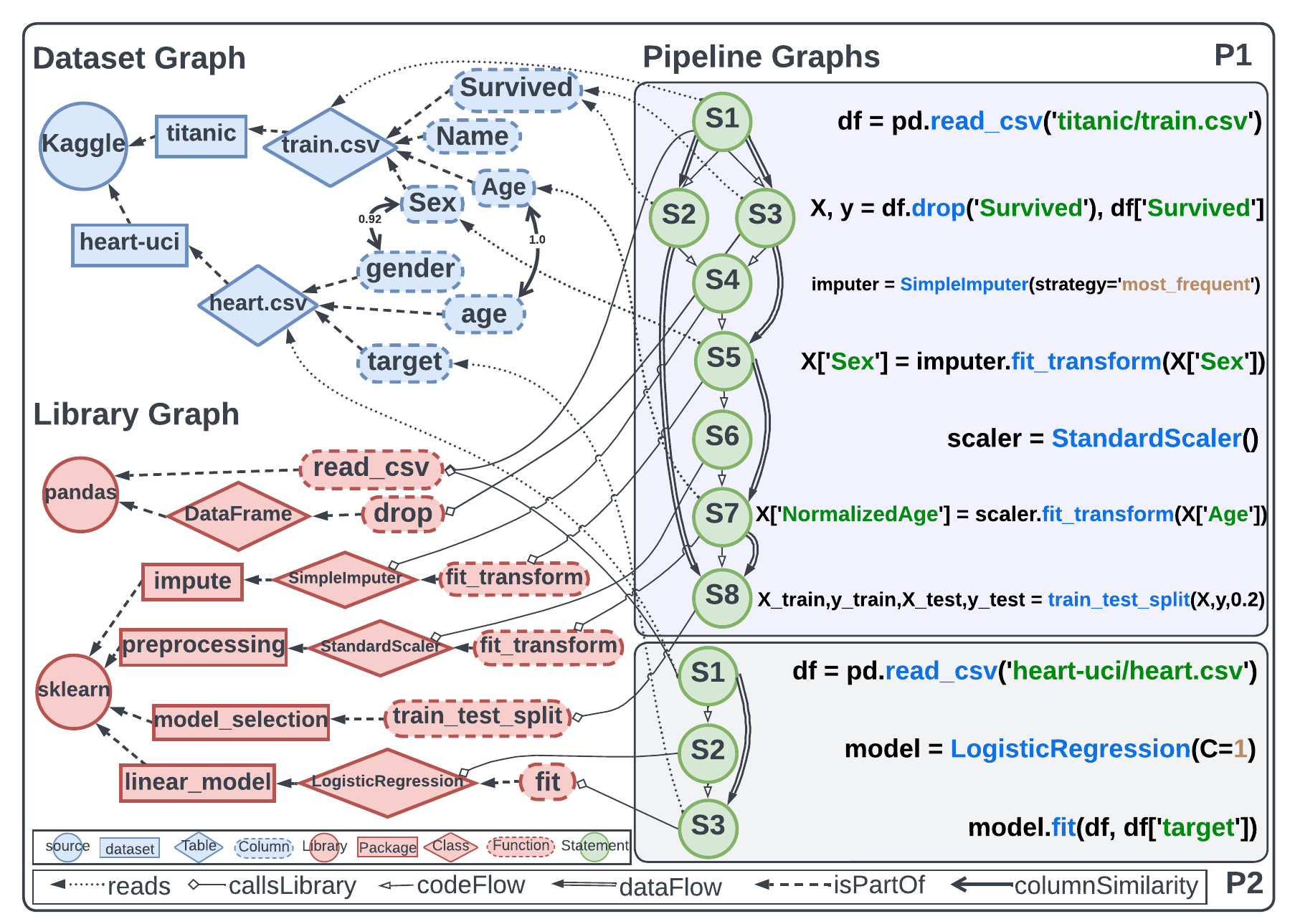} 
  \caption{An overview of the {\graphname} graph, which consists of the dataset, library, and pipeline graphs. Each pipeline is isolated in a named graph. Pipeline P1 is the abstraction of lines 2-10 in Figure \ref{fig:running_example}.
  }
    \label{fig:kglids_visualization}
\end{figure}

\nsstitle{The Linked Data Science (LiDS) ontology and graph. }
To store the created knowledge graph in a standardized and well-structured way, we developed an ontology for linked data science: the \emph{LiDS ontology}. 
Its main types of nodes (classes) are: datasets (with related nodes representing datasets, tables, and columns), libraries, and pipeline scripts (with related nodes describing statements). The \emph{LiDS ontology} conceptualizes the data, pipeline, and library entities in data science platforms, as illustrated in Figure~\ref{fig:kglids_visualization}. The ontology is specified in the Web Ontology Language (OWL 2) and has 13 classes, 19 object properties, and 22 data properties. OWL was chosen as a standard because of its integral support of interoperability and sharing data on the Web and across platforms. The URIs of classes and properties (relationships) have the prefix \emph{http://kglids.org/ontology/}, while data instances (resources) have the prefix \emph{http://kglids.org/resource/}.
%
We refer to the graph populating the ontology with instances of the classes as the \emph{LiDS graph}. Using the RDF standard~\cite{klyne2014rdf}, {\sysname} captures and describes the relationships among these entities and uses Uniform Resource Identifiers (URIs) for nodes and edges in the LiDS graph so that the graph can easily be published and shared on the Web. 
All entities are associated with an RDF label and RDF type to facilitate RDF reasoning on top of the {\graphname} graph. 

\subsection{The KGLiDS Interfaces and Storage}

The {\sysname} Interfaces is designed to meet the needs of data scientists in an open or enterprise setting whose objective is to derive insights from their datasets, construct pipelines, and share their results with other users. To achieve this goal, {\sysname} offers a number of interfaces (see component 3 in Figure~\ref{fig:overview}). Our interfaces include  a set of \emph{Pre-defined Operations}, \emph{Ad-hoc Queries} enabling users to query the {\graphname} graph directly. In addition, the \emph{Statistics Manager} helps collect and manage statistics about the system and the {\graphname} graph. Finally, the \emph{Model Manager}  enables data scientists to run analyses and train models directly on the {\graphname} graph to derive insights efficiently. Users can upload their models, explore the available ones, and use them in their applications.
Interoperability with other tools was one of {\sysname}'s design goals. Hence, {\sysname}, for instance, exports query results as Pandas DataFrame, a widely used format in data science~\cite{Petersohn21}. 
The {\sysname} portal supports access restrictions to prevent unauthorized users or could be public for anyone. Authorized users have access to query the LiDS graph or embeddings. However, accessing the actual data files in an enterprise may need another level of authorization.

\begin{figure}[t]
 \centering
  \includegraphics[width=\columnwidth]{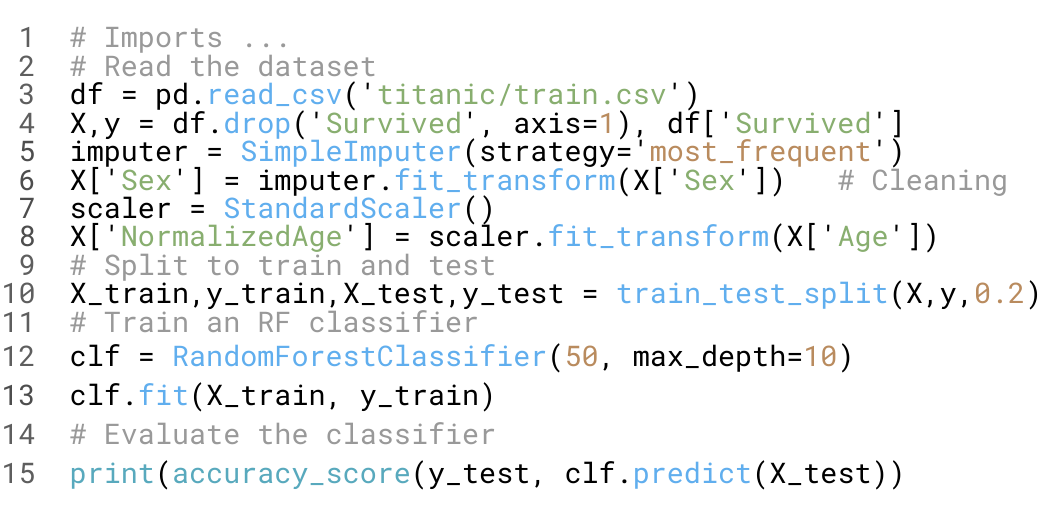} 
  \caption{A running example to demonstrate the {\sysname} Pipeline Abstraction. The pipeline loads a dataset using Pandas, performs data cleaning (imputation) and data transformation (scaling). The dataset is split into training and testing. Finally, a random forest classifier is fitted and evaluated.\shorten}
  \label{fig:running_example}
\end{figure}

{\sysname} maintains different types of information, namely the LiDS graph, the generated embeddings for columns and tables, and the machine learning models for different use cases. 
The current implementation of {\sysname} adopts the RDF model to manage the LiDS graph and uses GraphDB \cite{graphdb} as a storage engine. 
{\sysname} uses RDF-based knowledge graph technology because (i) it already includes the formalization of rules and metadata using a controlled vocabulary for the labels in the graphs ensuring interoperability~\cite{Villazon-TerrazasGRSRAP17,AMIE13}, (ii) it has built-in notions of modularity in the form of named graphs, for instance, each pipeline is abstracted in its own named graph~\cite{NamedGraphs2005}, (iii) it is schema-agnostic, allowing the platform to support reasoning and semantic manipulation, e.g., adding new labeled edges between equivalent artifacts, as the platform evolves~\cite{ZhengZPYSZ16,BursztynGMR15,GalarragaTHS15}, and (iv) it has a powerful query language (SPARQL) to support efficient query processing~\cite{lusail,MontoyaSH17,SchwarteHHSS11}.\shorten  

{\sysname} uses the RDF-star \cite{rdfstar} model, which supports annotating edges between nodes with metadata, which enables us, for instance, to capture the similarity scores for similarity edges between column nodes of datasets.
{\sysname} uses an embedding store, i.e., Faiss~\cite{faiss}, to index the generated embeddings and enable several methods for similarity search based on approximate nearest neighbour operations.
This allows users to query the LiDS graph based on the embeddings associated with graph nodes. Finally, {\sysname} also stores the ML models developed on top of the LiDS graph, e.g., GNN models for data cleaning and transformation.

\section{Semantic Abstraction}
\label{sec:governor}


The core component of the {\sysname} platform is the KG Governor, which captures the semantics of datasets, associated data science pipelines, and programming libraries to construct (i) pipeline graphs, e.g., each pipeline script is abstracted and stored in a separate named graph, (ii) a library graph for all programming libraries used by the abstracted pipelines, and (iii) a dataset graph, i.e., a global schema for datasets accessible by  {\sysname}. For pipeline and library graphs, {\sysname} performs static code analysis combined with documentation and dataset usage analysis to naturally interlink a pipeline graph into the dataset and library graphs. For the dataset graph, {\sysname} starts by profiling the datasets at the granularity of columns and then predicts links between columns.

\subsection{Pipeline Abstraction}
A data science pipeline, i.e., code or script, performs one or more data science tasks, e.g., data analysis, visualization, or modelling. The pipeline could be abstracted as a control flow graph, in which code statements are nodes and edges are a flow of instruction or data. 
The objective of pipeline abstraction is to have a language-independent representation of the pipeline semantics, such as code and data flow within a pipeline, invocations of built-in or third-party libraries, and parameters used in such invocations. This information can be obtained using dynamic or static program analysis.
While our pipeline abstraction is generalizable to several programming languages, the current implementation supports Python \cite{kaggle_survey}, which covers the vast majority of data science pipelines.

\subsubsection*{\textbf{Dynamic vs. Static Code Analysis}}
Dynamic analysis involves the execution of a pipeline and examining the memory traces of each statement at run time. Thus, it is more detailed and accurate but does not scale. This is because executing a pipeline is costly in terms of time and memory. It also involves setting up the runtime environment, which is not always possible due to, for example, deprecated libraries. In contrast, static code analysis is less accurate, especially for dynamic languages, such as Python and R. However, it scales well to thousands of pipelines as no pipeline execution is required. To overcome these limitations, {\sysname} combines static code analysis with library documentation and dataset usage analyses to have a rich semantic abstraction that captures the essential concepts in data science pipelines. 
Algorithm~\ref{alg:pipeline_abstraction} is the pseudocode of our Pipeline Abstraction algorithm. The main inputs of Algorithm~\ref{alg:pipeline_abstraction} are a set $S = \{s\}$, where $s$ is a pipeline script, $MD$ metadata of pipelines, such as information of datasets used in the pipeline, pipeline author, and its score, and $LD$ documentations of programming libraries. The algorithm also maintains the library graph for the used libraries and a named graph for each pipeline (lines 2 to 3). Our algorithm decomposes the pipeline abstraction into a set of independent jobs (line 5), where each job generates an abstracted graph (named graph) per a pipeline script (line 18).\shorten

\subsubsection*{\textbf{Static Code Analysis}}
Algorithm \ref{alg:pipeline_abstraction} applies static code analysis per $s$ (line 7). {\sysname} utilizes the lightweight static code analysis tools, which are natively supported by several programming languages, such as Python (via, for instance \texttt{ast} and \texttt{astor}) or R (via, for instance, CodeDepends). Each statement corresponds to a variable assignment or a method call. If there are multiple calls in a single line, they will correspond to multiple statements. For each statement, we store the following: i) \emph{code flow}, i.e., the order of execution of statements, ii) \emph{data flow}, i.e., subsequent statements that read or manipulate the same data variable, and iii) Control flow type, i.e., whether the statement occurs in a loop, a conditional, an import, or a user-defined function block, which captures the semantics of how a statement is executed, and iv) statement text, i.e., the raw text of the statement as it appears in the pipeline. We discard from our analysis statements that have no significance in the pipeline semantics, such as \texttt{print()}, \texttt{DataFrame.head()}, and \texttt{summary()}.\shorten  

\begin{algorithm}[t]
\SetKwFor{For}{for}{}{}
\SetKwIF{If}{ElseIf}{Else}{if}{}{else if}{else}{end if}
\DontPrintSemicolon
\footnotesize
\caption{Pipeline Abstraction} 
\label{alg:pipeline_abstraction}
\SetKwInput{Input}{Input}
\Input{Pipeline Scripts $\mathcal{S}$, Pipeline Metadata $\mathcal{MD}$,\\ \qquad\quad\  Programming Library Documentation $\mathcal{LD}$} 

\SetKwComment{Comment}{$\triangleright$\ }{}

\SetKwProg{Main}{Main Node}{:}{}
\Main{}{

library\_hierarchy $\gets$ build\_library\_hierarchy\_subgraph($\mathcal{LD}$) \\
pipeline\_metadata $\gets$build\_pipeline\_metadata\_subgraph($\mathcal{MD}$) \\
json.dump(library\_hierarchy, pipeline\_metadata)\\
$\mathcal{S}_{rdd}.map(analyze\_pipeline\_script)$
}
\BlankLine

\SetKwProg{Worker}{Worker (Parallel)}{:}{}
\Worker{analyze\_pipeline\_script(s)}{
    g $\gets$ static\_code\_analysis(s) \Comment*[r]{ Control and data flows}
    \For{node $\in$ g \qquad\qquad\qquad\qquad}{
    \Comment*[r]{ Documentation Analysis} 
    \If{node calls library \texttt{lib} $\in$ $\mathcal{LD}$}{
             node.return\_type $\gets$ $\mathcal{LD}$ [\texttt{lib}].return\_type \\
             node.parameter\_names$\gets$$\mathcal{LD}$[\texttt{lib}].parameters.names \\
             \For{\texttt{p} $\in$ $\mathcal{LD}$ [\texttt{lib}].parameters and \texttt{p} $\notin$ \textrm{node.parameters}}{
                node.default\_parameters += (\texttt{p}.name, \texttt{p}.value)
             }

          }
            
              \Comment*[r]{ Dataset Usage Analysis} 
             \If{node calls \texttt{pandas.read\_csv(dataset\_name.csv)}}{
                node.detected\_dataset\_read $\gets$ \texttt{dataset\_name}
             }
            \If{node calls \texttt{pandas.DataFrame[column\_name]}}{
                node.detected\_column\_reads += \texttt{column\_name}
            }
    }

    json.dump(g) \Comment*[r]{ Save abstracted pipeline graph}
} 
\end{algorithm}

\subsubsection*{\textbf{Documentation Analysis}}
Static analysis is not sufficient to capture all semantics of a pipeline. For instance, it cannot detect that \texttt{pd.read\_csv()} in line 3 in Figure \ref{fig:running_example} returns a Pandas DataFrame object. In Algorithm~\ref{alg:pipeline_abstraction}, we enrich the static program analysis using the documentation of data science libraries (lines 9 to 13). Each statement from static program analysis is enriched by the library it calls, names and values of parameters, including implicit and default ones, and data types of return variables. For each class and method in the documentation, we build a JSON document containing the names, values, and data types of input parameters, including default parameters, as well as their return data types. This analysis enables accurate data type detection for library calls. It also allows the inference of names of implicit call parameters, such as \texttt{n\_estimators}, the first parameter in line 12 in Figure \ref{fig:running_example}. A useful by-product of documentation analysis is the library graph, indicating methods belonging to classes, sub-packages, etc. (shown in red in Figure \ref{fig:kglids_visualization}). This is useful for deriving exciting insights related to data science programming languages. For example, it helps find which libraries are used more frequently than others. {\sysname} enables retrieving this kind of insight via queries against the LiDS graph.

\subsubsection*{\textbf{Predicting Dataset Usage and Graph Linker}}
The critical aspect of our LiDS graph is the realization of connections between pipeline statements and the tables or columns used by the pipeline.  These connections enable the novel use-cases of linked data science platforms.
In {\sysname}, we build these links in two phases. First, 
Algorithm~\ref{alg:pipeline_abstraction} applies dataset usage analysis to predict such cases and adds a node of the \textit{Predicted Dataset Usage} (lines 14 to 17). If a statement reads a table via \texttt{pandas.read\_csv} (e.g. line 3 in Figure \ref{fig:running_example}) or a column via string indices over DataFrames (e.g. line 6 in Figure \ref{fig:running_example}), such tables or columns are predicted as potential reads of actual data. Second, because not all predicted columns or tables exist in the raw dataset, these predicted nodes are verified by the Graph Linker against the \texttt{Data Global Schema} of the corresponding dataset when constructing the knowledge graph. For instance, the \texttt{NormalizedAge} column in Line 8 of Figure \ref{fig:running_example} is a user-defined column; it is predicted as a column node in the first phase and is then removed after matching against the dataset graph, see Figure~\ref{fig:kglids_visualization}. 

\subsection{Our Embedding-based Data Profiling}
\label{sec:data_profiling}
{\sysname} profiles datasets to learn representations (embeddings) of columns and tables, then generates fixed-size and dense embeddings based on their content (e.g., column values) and semantics (e.g., table or column names). Moreover, our profiler collects statistics and classifies columns into 7 fine-grained types using our data type inference module. Inspired by~\cite{MuellerS19}, we developed a deep learning model to generate column learned representations (\emph{CoLR}) based on their content. For embeddings based on label semantics, i.e., column names, we developed a method based on Word Embeddings~\cite{semantic_model}. {\sysname} analyzes datasets at the level of individual columns. 
We developed {\sysname} to profile datasets at scale. This is achieved through two main steps: \myNum{i} using CoLR to get fixed-size embeddings per column and \myNum{ii} performing a pairwise comparison between columns of the same type. Our data profiling is developed using PySpark to enable distributed computation.
Moreover, {\sysname} handles files of different formats, such as CSV and JSON, and connects to relational DB and NoSQL systems. 
Algorithm~\ref{alg:data_profiler} is the pseudocode of our data profiling algorithm. 
It receives: \myNum{i} a dataset $D$ consisting of one or more tables, \myNum{ii} a set of CoLR models $\mathcal{H}_{\theta, \mathcal{T}}$ to generate embeddings of columns the fine-grained data types, and \myNum{iii} an NER model and a set of word embeddings and to predict fine-grained types. In Algorithm~\ref{alg:data_profiler}, tables are broken down into a set of columns. Then, {\sysname} uses Spark to generate a column profile (JSON) per column (lines 2 and 3).


\begin{algorithm}[t]
\DontPrintSemicolon
\footnotesize
\caption{Data Profiling} 
\label{alg:data_profiler}
\SetKwInput{Input}{Input}

\Input{Datasets $\mathcal{D}$, CoLR Models $\mathcal{H}_{\theta, \mathcal{T}}$, Word Embeddings $\mathcal{W}$, Pre-trained NER Model $f_{\sigma}$} 

\SetKwComment{Comment}{$\triangleright$\ }{}

\SetKwProg{Main}{Main Node}{:}{}
\Main{}{

$\textrm{columns}_{rdd} = \{c ; c \in t; t \in \mathcal{D}\}$ \Comment*[r]{ Columns in all tables}
$\textrm{columns}_{rdd}.map(profile\_column)$
}

\BlankLine

\SetKwProg{Worker}{Worker (Parallel)}{:}{}
\Worker{profile\_column(col)}{
    
    $\mathit{M} = \textrm{col.metadata}$ \Comment*[r]{ Table and dataset membership}
    $ \mathit{fgt} = \textrm{infer\_fine\_grained\_type}(col, \mathcal{W}, f_{\sigma})$ \\
    $\mathit{S} = \textrm{collect\_stats} (col, \mathit{fgt})$ \Comment*[r]{ Statistics e.g. \#NaNs}
    $\mathit{E} = [0]^{300\times1}$ \Comment*[r]{ 300-Dimensional column embedding} 
    \For{val $\in$ col.sample(max($0.1|col|, 1000$))}{
        $\mathit{E} \gets \mathit{E} + \frac{1}{|col|} h_{\theta,\mathit{fgt}}(val)$ \Comment*[r]{ Avg. CoLR over values}
    }
    
    $\mathcal{CP} = \{\mathit{M}, \mathit{fgt}, \mathit{S}, \mathit{E}\}$ \Comment*[r]{ Column profile}
    json.dump($\mathcal{CP}$) \Comment*[r]{ Store}

}

\end{algorithm}






\subsubsection*{\textbf{Data Type Inference}} 
{\sysname} predicts similarities between columns across tables by performing a pairwise comparison between embeddings of columns having the same fine-grained type -- this helps reduce the cost of constructing the dataset graph. {\sysname} infers for each column (line 6) a fine-grained data type out of 7  types, namely: integers, floats, booleans, dates, named entities (e.g. names of persons, locations, languages), natural language texts (e.g. product reviews, comments), and generic strings that do not fall into the previous categories (e.g. postal codes or IDs). Named entities are predicted using a pre-trained named entity recognition (NER) model \cite{ner_model} trained on the OntoNotes 5 \cite{ontonotes} dataset, which recognizes 18 entity types including persons, countries, organizations, products, and events. In addition, natural language texts are predicted based on the existence of corresponding word embeddings for the tokens.
Our fine-grained types drastically cut false positives in column similarity prediction by limiting comparisons to columns of the same type.\shorten

\subsubsection*{\textbf{Dataset Embeddings}}
\label{sec:dataset_embeddings}
The data profiling component generates a column learned representation (CoLR) for each column based on its fine-grained type and actual values using our pre-trained embedding models. The CoLR models capture similarities between column values and provide three main advantages to KGLiDS. First, higher accuracy of predicted column content similarities in contrast to hand-crafted meta-features, which have been shown to fail when a column distribution does not match the designed features \cite{MuellerS19,sherlock}. Second, enabling data discovery without exposing datasets' raw content is invaluable in an enterprise setting, where access to the raw data might be restricted. Third, a compact representation of fixed-size embeddings, regardless of the actual dataset size, greatly reduces the storage and memory requirements. 

Two columns have similar embeddings if their raw values have high overlap, have similar distributions, or measure the same variable -- even with different distributions (e.g. \texttt{area\_sq\_ft} is similar to \texttt{area\_sq\_m}). To generate a column embedding, a 10\% random sample is taken from the column, and a neural network $h_{\theta}$ is applied to each value and averaged for the entire sample (lines 8-10). We trained $h_{\theta}$ on a collection of 5,500 tables from Kaggle \cite{kaggle} and OpenML \cite{OpenML}. The input is column pairs, predicting similarity (binary target) with binary cross-entropy loss \cite{cross_entropy}.
In KGLiDS, the embedding of a table is the concatenation of aggregated column embeddings per fine-grained data type:\shorten 

\begin{equation}
\label{equation_dataset_embedding_ref}
    h_{\theta}(\mathcal{D}) = \concat_{fgt \in \mathcal{T}} \frac{1}{|c_{fgt}|} \sum_{c_{fgt} \in \mathcal{D}}{h_{\theta, fgt}(c_{fgt})}
\end{equation}
where $|c_{fgt}|$ is the number of columns in $\mathcal{D}$ with the fine-grained type $fgt$. Similarly, an embedding of a dataset is an aggregation of its tables' embeddings. The Data Profiling stores the generated embeddings in the embedding store. For simplicity, we do not show the generation of the table's embeddings. Algorithm~\ref{alg:data_profiler} generates a column's profile containing the predicted fine-grained type $\textit{fgt}$, the generated embeddings $E$, the column statistics $S$, and metadata $\mathcal{M}$  and dumps it as a JSON document (lines 11 and 12). Algorithm~\ref{alg:data_profiler} is designed to work with independent tasks at two levels. First, the dataset is decomposed into independent tables. Second, each table is decomposed into a set of columns, where most computations are done. This design profiles datasets at scale.

\subsection{The {\graphname} Graph Construction}
\label{subsec:dataset_graph}
\begin{algorithm}[t]
\SetKwIF{If}{ElseIf}{Else}{if}{}{else if}{else:}{end if}
\footnotesize
\caption{Data Global Schema Builder} 
\label{alg:data_items_subgraph}
\SetKwInput{Input}{Input}
\Input{Column Profiles $\mathcal{CP}$, Similarity Thresholds: $\alpha, \beta, \theta$} 
\SetKwComment{Comment}{$\triangleright$\ }{}
\DontPrintSemicolon

\SetKwProg{Main}{Main Node}{:}{}
\Main{}{
$\mathcal{CP}_{rdd}$.mapPartitions(\texttt{column\_metadata\_worker})
}
\BlankLine

\SetKwProg{Worker}{Worker (Parallel)}{:}{}
\Worker{column\_metadata\_worker(cp)}{
g = \texttt{create\_metadata\_subgraph($cp$)} \\
json.dump(g) \Comment*[r]{Save metadata subgraph}
} 
\BlankLine

\Main{}{
\Comment*[l]{column pairs with the same fine-grained type}
$\mathcal{P} = \{(cp_i, cp_j) \ | \ cp_i,cp_j \in \mathcal{CP}; i \neq j; \mathcal{T}_{cp_i} = \mathcal{T}_{cp_j} \} $ \\

$\mathcal{P}_{rdd}$.map(\texttt{column\_similarity\_worker}) \\
}
\BlankLine

\Worker{column\_similarity\_worker($cp_i, cp_j$)}{
    $g = \phi$ \\
    \If{word\_embed\_sim$(cp_i.label, cp_j.label)  \ge \alpha$ :}{
        add\_edge(g, $cp_i, cp_j$, "LabelSimilarity", $\alpha$)
    }
    \If{$\mathcal{T}_{cp_i}$ == "boolean" :}{
        \If{$(1 - \mid cp_i.true\_ratio - cp_j.true\_ratio\mid)\ \ge \beta$ :}{
            add\_edge(g, $cp_i, cp_j$, "ContentSimilarity", $\beta$)
        }
    }
    \Else{
        \If{$colr\_embed\_sim(cp_i.embed, cp_j.embed)\ \ge \theta$ :}{
            add\_edge(g, $cp_i, cp_j$, "ContentSimilarity", $\theta$)
        }
    }
    
    json.dump(g) \Comment*[r]{Save column similarity subgraph}
}
\BlankLine

\Main{}{
    $G=\phi$ \Comment*[r]{Data global schema graph}
    \For{subgraph g}{
        $G \gets G \cup g$
    }
    \Return{G} 
}

\end{algorithm}

This section highlights the dataset graph construction and interlinking it with the pipeline graphs. The {\graphname} graph is maintained as a Web-accessible graph based on our ontology.

\subsubsection*{\textbf{Data Global Schema}}
Algorithm~\ref{alg:data_items_subgraph} illustrates the pseudocode of our algorithm to construct the dataset graph. The algorithm receives a set of column profiles $\mathcal{CP}$ and a set of similarity thresholds. Each profile contains the predicted fine-grained type $\mathcal{T}$, the generated embeddings $E$, the column statistics $S$, and metadata $\mathcal{M}$, which are generated by Algorithm~\ref{alg:data_profiler}. First, Algorithm~\ref{alg:data_items_subgraph} constructs a metadata subgraph, which contains the hierarchy structure of the datasets and statistics collected for each column. Next, the similarity relationships are checked between all possible column pairs having the same fine-grained data type and exist in different tables. Algorithm~\ref{alg:data_items_subgraph} distributes the processing of the pairwise comparisons in a MapReduce fashion (lines 9 to 19). 

Each worker takes a pair of profiles and generates the similarity edges, i.e., predicates, between them. Two columns have similarity relationships if they have higher similarity scores than the predefined thresholds for the following similarities: \myNum{i} \textbf{label similarity}:  exists between columns that have similar column names based on GloVe Word embeddings \cite{glove} and a semantic similarity technique \cite{semantic_model} (lines 11-12). \myNum{ii} \textbf{content similarity} exists between columns that have similar raw values. For all fine-grained types except booleans, content similarity is based on the cosine distance between their column embeddings (lines 16-18), while for booleans, it is based on the difference in \textit{true ratio}, i.e., the percentage of values that equal \texttt{True} (lines 13-15).
The thresholds, $\alpha$, $\beta$, and $\theta$ are user-defined and control the similarity thresholds above which column similarity relationships are materialized in the LiDS graph. High similarity thresholds result in less but more accurate edges, i.e. high precision and low recall. Conversely, lower similarity thresholds might be used when high recall is desirable.
Finally, the dataset graph is constructed (lines 21 to 24). 
{\sysname} utilizes the predicted relationships between columns to identify unionable and joinable tables. Two tables are unionable or joinable if one or more of their columns have high label or content similarity relationships, respectively.  The similarity score between two tables is based on both the number of similar columns and the similarity scores between them.\shorten

\section{Data Science On-Demand Automation}
\label{sec:applications}

Existing data cleaning and transformation systems analyze the raw dataset to recover the original missing values or find the optimal normalization. Hence, they do not scale and are time-consuming. Instead of working at the raw data level, KGLiDS trains GNN models for data cleaning and transformation using embeddings of datasets and sets of operations applied to them to impute missing values or normalize the data. Hence, our GNN models can be used on-demand to provide interactive recommendations.
Our on-demand automation is based on GNN models trained to predict a near-optimal operation/estimator for an unseen dataset $d_u$ based on the set of operations/estimators used with the most similar dataset $d_s$ in our KG. We measure the cosine similarity between the embeddings of $d_u$ and  $d_s$. Each $d_u$ is associated with a modelling task, such as predicting survivals in the Titanic dataset shown in Figure 3. For example, existing data cleaning systems perform general cleaning to recover the original data. In contrast to these systems, our on-demand automation aims to maximize the model performance for the associated task.\shorten

\subsection{Our GNN Training and Inference}
\sysname{} could be queried to fetch the cleaning or transformation operations and dataset nodes of type columns or tables used as input. For instance, the training dataset could include frequently used transformation operations to train data transformation recommendation models. Such cases can be observed in figure \ref{fig:kglids_visualization}, where the operation StandardScaler is applied to the column 'Age'. Then, we can train a GNN model on the extracted subgraph of these operations and associated columns. We use CoLR embeddings to initialize the embeddings of dataset nodes, i.e., nodes of type columns or tables. 
We trained \sysname{}'s models on top of a KG constructed from top-rated 1000 Kaggle datasets and 13800 pipeline scripts with the highest number of votes. 
We utilized GraphSAINT~\cite{GraphSAINT} to train GNN models for our node classification models for predicting operations for data cleaning and transformation.

In the inference phase,  the GNN model takes the unseen dataset in the form of a DataFrame and calculates the CoLR embedding for each column. It then leverages a task-specific pre-trained model to predict the most relevant operation to be applied to the dataset. Our GNN models are based on 
fixed-size embeddings that enable concise summarization of a dataset of large content. Hence, our on-demand automation for data preparation scales to large datasets. 

\sysname{} provides several task-specific APIs such as:\shorten
\begin{Verbatim}[commandchars=\\\{\}]
{\small
\textbf{{\color{blue}cleaning_ops}} = \textbf{recommend_cleaning_operations(}df\textbf{)}
}
\end{Verbatim}
that provide data cleaning recommendations. Additionally, \sysname{} provides another set of APIs that enable data scientists to directly select and apply the recommended data cleaning operations to their dataset without requiring explicit code. For instance, the following API:\shorten
\begin{Verbatim}[commandchars=\\\{\}]
{\small
\textbf{apply_cleaning_operations(}\textbf{{\color{blue}cleaning_ops}},df\textbf{)}
}
\end{Verbatim}
takes the base dataframe and selected cleaning operation as input and returns the cleaned dataset as output. \sysname{} has similar APIs to support data transformation and modelling.  

\subsection{Our GNN Models for Data Cleaning}
\label{subsec:data_cleaning}

We formalize the GNN task of data cleaning as a multiclass node classification problem. Given a dataset $\mathcal{D}$ with missing values $\mathcal{M}=\{m_1,m_2,...,m_n\}$ and a machine learning task $\mathcal{L}$, the recommendation task is to predict a near-optimal cleaning operation to handle $\mathcal{M}$ such that the model performance of $\mathcal{L}$ is improved. 
Our model predicts the near-optimal cleaning operation from a set of operations applied by other data scientists in other datasets similar to $\mathcal{D}$.\shorten 

Our GNN model is initialized using table embeddings derived from our column embeddings, which are calculated by averaging the embeddings of the columns in the table that contain missing values. Separate averages are computed for each column datatype, and these averaged embeddings are then concatenated. 
The embeddings used to initialize the GNN model are of length 1800, which is the concatenation of embeddings for six fine-grained column types. All columns of the same type are aggregated together in embeddings of size 300.   
The output of the model can be one of 5 cleaning operations (Fillna, Interpolate, SimpleImputer, KNNImputer, IterativeImputer). The GNN model has one layer, as there is only one edge between a given table and its cleaning operation.\shorten

\subsection{Our GNN Models for Data Transformation}
\label{subsec:data_transformation}

Similar to data cleaning, we formalize the GNN task of data transformation recommendation as a classification problem. Given a dataset $\mathcal{D}$ with features $\mathcal{F}=\{f_1,f_2,...,f_n\}$ and a machine learning task $\mathcal{L}$, the recommendation task is to predict a set of transformations $\mathcal{T}=\{t_1,t_2,...,t_m\}$ to improve the model performance of $\mathcal{L}$ based on the transformation techniques applied on similar datasets to $\mathcal{D}$. The problem of data transformation recommendation is subdivided into two primary steps: 1) recommending scaling transformations (applied to the entire dataset) and 2) recommending unary feature transformations (applied to an individual feature of the dataset). The primary motivation behind employing scaling transformations prior to unary transformations lies in addressing the challenges of varying data magnitudes \cite{scaling_helps}. By scaling the data, we ensure that all features are treated on an equal footing to prevent certain features from overshadowing others due to their scale. \shorten

We support two types of transformation: Table transformations (Standard Scaler, Minmax Scaler, and Robust Scaler) and column transformations (log and sqrt). Each has its own training dataset and GNN model. 
The table transformation model is trained using averaged embeddings of each column type concatenated (1800) as initialization. The column transformation did not require any aggregation in its initialization, as each column was directly associated with its embedding of size 300.\shorten

\subsection{Our AutoML Support using GNNs}
The AutoML problem aims to select the near-optimal ML estimator, i.e., classifier or regressor, with the set of parameters for a certain dataset. KGpip~\cite{kgpip} is the state-of-the-art AutoML system based on GNNs trained on top of a data science KG. The initial version of KGpip was trained using a KG generated using a general-purpose code abstraction tool called GraphGen4Code\cite{graph4code}. Hence, the offline training phase of KGpip involves extensive graph filtration to remove noisy nodes that are not relevant to data science artifacts, as stated in~\cite{kgpip}. The inference phase includes using the KGpip model to predict a classifier. Then, KGpip utilizes a hyperparameter optimizer to search for near-optimal set of parameters.

{\sysname} improves the KGpip system in two ways. First, the {\graphname} graph maintains only the semantics of data science artifacts. Hence, {\sysname} helps KGpip avoid the extensive graph filtration. Second and more importantly, the {\graphname} graph augments all function calls with the parameter names used by analyzing the relevant programming documentation,  including implicit (nameless) and default (unspecified) parameters). For AutoML, function parameters are important because hyperparameters of ML models are specified as function parameters. Hence, the {\graphname} graph includes for each ML model the set of pairs (hyperparameter name and value) used. The KG constructed using GraphGen4Code does not include this information. We improved the KGpip inference pipeline by utilizing this information to prune the hyperparameter search space.  This revised pipeline significantly reduces the search time and leads to a significant improvement in the accuracy of the KGpip AutoML system.\shorten 


\section{The {\sysname} Interfaces}
\label{sec:interfaces}

The primary users of {\sysname} are data scientists in an open or enterprise setting whose objective is to derive insights from their datasets, construct pipelines, and share their results with other users. To achieve this goal, {\sysname} offers a number of interfaces (see component 3 in Figure~\ref{fig:overview}). Our interfaces include  a set of \emph{Pre-defined Operations} and \emph{Ad-hoc Queries}, enabling users to query the raw {\graphname} graph directly. In addition, the \emph{Statistics Manager} helps collect and manage statistics about the system and the {\graphname} graph. Finally, the \emph{Model Manager}  enables data scientists to run analyses and train models directly on the {\graphname} graph to derive insights. 

We developed the {\sysname} Interfaces library as a Python package that provides simple API interfaces\shorten\footnote{A Colab notebook demonstrating the {\sysname} APIs is available at \url{https://colab.research.google.com/drive/1dDiGh1KwJibR2pVjiMXFpHIHxfgByYxZ}}, allowing users to directly access the {\sysname}  storage. We designed these APIs to formulate the query results as a Pandas Dataframe, which Python libraries widely support. Thus, data scientists can use our APIs interactively or programmatically while writing their pipeline scripts via Jupyter Notebook or any Python-based data science platform. Due to limited space, the remainder of this section focuses on pre-defined operations.

Let us consider a scenario where a data scientist is interested in predicting heart failure in patients and illustrate how {\sysname}'s pre-defined operations can help achieve this goal. 

\nsstitle{Search Tables Based on Specific Columns.}
To get started, the data scientist would like to find relevant datasets using keyword search, the following operation supports this: 
\begin{Verbatim}[commandchars=\\\{\}]
{\small
  \textbf{{\color{blue}table_info} = search_keywords}([['heart', 
}
\end{Verbatim}
{\small
\begin{verbatim}
        'disease'], 'patients']])
\end{verbatim}
}
{\sysname} will then perform a search in {\graphname} using the conditions that are passed by the user who has the possibility to express conjunctive (\texttt{AND}) and disjunctive conditions (\texttt{OR}) using nested lists. 
In the example above `heart' and `disease' are conjunctive and `patients' is a disjunctive condition. 
Let us assume the data scientist has found the following two datasets of interest:
\texttt{heart-failure-prediction}, \texttt{heart-failure-clinical-data}. 

\nsstitle{Discover Unionable Columns.}
In the next step, let us assume that the data scientist would like to combine the two tables into one. Since it is very unlikely that the two tables in our example come with exactly the same schema, the data scientist is seeking assistance to identify matching (unionable) columns expressing the same information. 
{\sysname} provides support to automatically recommend a schema for the merged table containing all columns from both input tables that can be matched (unionable columns). The output will be a Pandas DataFrame. This can be expressed as:
\begin{Verbatim}[commandchars=\\\{\}]
{\small
  \textbf{find\_unionable\_columns}({\color{blue}\textbf{table_info}}.iloc[0], 
}
\end{Verbatim}
{\small
\begin{Verbatim}[commandchars=\\\{\}]
        {\color{blue}\textbf{table_info}}.iloc[1])
\end{Verbatim}
}
\nsstitle{Join Path Discovery.}
Let us now assume that the data scientist would like to join the obtained table with a table from another dataset to obtain a richer set of features for downstream tasks. Let us further assume that the table is not directly joinable.
Hence, {\sysname} suggests intermediate tables (restricted to a join path with 2 hops in our example) and displays the potential join paths. 
This is done by computing an embedding of the given DataFrame (\texttt{df}), finding the most similar table in the {\graphname} graph, and determining potential join paths to the given target table. This could be expressed as follows:\shorten 
\begin{Verbatim}[commandchars=\\\{\}]
{\small
  \textbf{get_path_to_table}({\color{blue}\textbf{table_info}}.iloc[0],hops=2) 
}
\end{Verbatim}

{\sysname} also supports more challenging variations, such as identifying the shortest path between two given tables.


\nsstitle{Library Discovery.}
Before switching to the machine learning phase of their pipeline, the data scientist would like to have a look at the most used libraries by their fellow users. The library graph of {\sysname} can be utilized to retrieve the number of unique pipelines calling a specific library. This can be expressed as:\shorten

\begin{figure}[t]
\ncp\ncp
 \centering
  \includegraphics[width=.97\columnwidth]{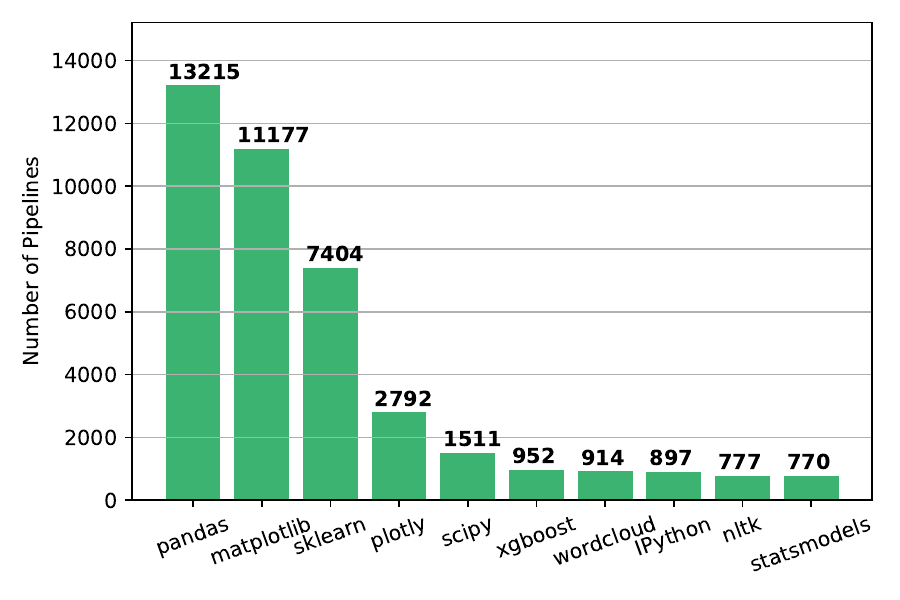} 
  \ncp\ncp\ncp\ncp\ncp\ncp
  \caption{Top 10 libraries used in 13k Kaggle pipelines.}
  \label{fig:library_stats}
  \ncp\ncp\ncp\ncp\ncp\ncp
\end{figure}

\begin{Verbatim}[commandchars=\\\{\}]
{\small
  \textbf{get_top_k_library_used}(k: int)
}
\end{Verbatim}

where $k$ is the number of libraries. {\sysname} also plots the corresponding bar chart with the top-k used libraries in all the pipelines (see Figure \ref{fig:library_stats}).  Data scientists can quickly get statistics about different data science artifacts, i.e., used libraries.
As the data scientist wishes to predict heart failure and is not an expert in building ML pipelines, {\sysname} assists the data scientist by providing a way to become familiar with the most used libraries in ML pipelines for a specific task, classification in this case. 
This can be expressed as follows:\shorten 
\begin{Verbatim}[commandchars=\\\{\}]
{\small
    \textbf{get_top_used_libraries}(k=10,
}
\end{Verbatim}
\begin{Verbatim}[commandchars=\\\{\}]
{\small
        task='classification')
}
\end{Verbatim}

\nsstitle{Pipeline Discovery.}
After reading more about some of the most used libraries (namely, Pandas, Scikit-learn, and XGBoost), the data scientist is interested to see example pipelines where the following components are used: \texttt{pandas.read\_csv}, \texttt{XGBoost.XGBClassifier}, and \texttt{sklearn.f1\_score}.
This can be expressed as: 
\begin{Verbatim}[commandchars=\\\{\}]
{\small
  \textbf{get_pipelines_calling_libraries}(
}
{\small
        'pandas.read_csv', 
}
{\small
        'xgboost.XGBClassifier',
   
}
{\small
        'sklearn.metrics.f1_score')
}
\end{Verbatim}
{\sysname} returns a DataFrame containing a list of pipelines matching the criteria along with other important metadata. 


\nsstitle{Transformation Recommendation.}
Before training a model, its important to perform the necessary pre-processing and transformations. For the same {\sysname} offers a function to recommend transformations for a given dataset. The data scientist in our example can make use of this function as follows:\shorten 
\begin{Verbatim}[commandchars=\\\{\}]
{\small
    \textbf{recommend_transformations}( 
}
{\small
        dataset='heart-failure-prediction')  
}
\end{Verbatim}
\sysname{} returns the set of transformations that could be applied on the given dataset such as sklearn's MinMaxScaler, OneHotEncoder etc. Based on these recommendations the data scientist can then easily transform their data to generate a much more representative dataset for model building.

\nsstitle{Classifier Recommendation.}
Afterwards, the data scientist needs to decide which classification model to use and would like to retrieve suggestions:
\begin{Verbatim}[commandchars=\\\{\}]
{\small
    {\color{blue}\textbf{model_info}} \textbf{= recommend_ml_models}(
}
\end{Verbatim}
\begin{Verbatim}[commandchars=\\\{\}]
{\small
        dataset='heart-failure-prediction', 
}
{\small
        task='classification')
}
\end{Verbatim}

{\sysname{}} returns a dataframe with the list of all classifiers that have been used for the given dataset along with their score to assist the data scientist in finalizing the model training.\shorten 

\nsstitle{Hyperparameter Recommendation.}
In the final step, the data scientist would like {\sysname} to provide help with finding a promising configuration for the hyperparameters of the chosen classifier. This can be expressed as follows: 
\begin{Verbatim}[commandchars=\\\{\}]
{\small
    \textbf{recommend_hyperparameters}(}{\small{\color{blue}\textbf{model_info}}.iloc[0])
}
\end{Verbatim}

The data scientist can use the configuration values of \textbf{recommend\_hyperparameters} to train a model based on the seen ones.  This enables hyperparameter optimization with ease based on the best practices adopted in thousands of pipelines.

\section{Experimental Evaluation}
\label{sec:evaluation}
We evaluate the performance of {\sysname}'s components against the state-of-the-art (SOTA) systems in data discovery, semantic abstraction of code, data cleaning, data transformation, and AutoML. Unlike {\sysname}, none of these systems provides simultaneous support for all these functionalities.

\begin{table}[t]
\vspace*{-2ex}
\caption{Data Discovery Benchmarks. The breakdown of column data types is obtained using our data profiler.}
\vspace*{-2ex}
\label{tab:data_discovery_dataset_stats}
\setlength\tabcolsep{5pt} 
\begin{tabular}{lllll}
\toprule
\textbf{Statistic}                         & \textbf{\begin{tabular}[c]{@{}l@{}}$\mathbf{D^3L}$~\\\texttt{Small}\end{tabular}} & \textbf{\begin{tabular}[c]{@{}l@{}}\texttt{TUS}~\\\texttt{Small}\end{tabular}} & \textbf{\begin{tabular}[c]{@{}l@{}}\texttt{SANTOS}~\\\texttt{Small}\end{tabular}} & \textbf{\begin{tabular}[c]{@{}l@{}}\texttt{SANTOS}~\\\texttt{Large}\end{tabular}} \\
\midrule
Size (GB)                         & 1.3       &  1.2     & 0.4        & 11.5         \\
No. tables              & 654       &  1,530   & 550        & 11,090       \\
No. query tables                  & 50        & 150      & 50         & 80           \\
Avg. No. unionable tables         & 110       &  163     & 14         & -            \\
Avg. No. rows per table           & 12,207    &  4,457   & 6,921      &  7,718        \\
Total columns                     & 8,767     &  14,810  & 6,336      & 121,796      \\
\midrule
\texttt{int} cols.              & 1,885       &  1,222   & 1,267      & 25,618       \\
\texttt{float} cols.            & 513         &  288     & 271        & 5,702        \\
\texttt{boolean} cols.          & 8           &  111     & 110        & 1,173        \\
\texttt{date} cols.            & 661          &  884     & 331        & 6,891        \\ 
\texttt{named\_entity} cols.    & 516         &  1,766   & 1,053      & 18,897       \\
\texttt{natual\_language} cols. & 4,241       &  9,345   & 2,908      & 53,502      \\
\texttt{string} cols.           & 957         &  1,194   & 396        & 10,013      \\
\bottomrule
\end{tabular}
\vspace*{-3ex}
\end{table}

\nsstitle{Experimental Setup.}
We used GraphDB\footnote{\url{https://www.ontotext.com/products/graphdb/}} version \textit{10.4.1} as an RDF engine to store the {\graphname} graph. We used two different hardware settings for our experiments. For the semantic abstraction of code, data cleaning, data transformation, and AutoML experiments, we used Ubuntu 22.04, a 16-core CPU at 2.40 GHz, and 189 GB of RAM. For the data discovery experiments, we used Ubuntu 22.04, a 64-core CPU at 2.45 GHz, 1TB of RAM, and an Nvidia A100 GPU.\shorten

\subsection{Data Discovery Systems}

We evaluate {\sysname} against two SOTA data discovery systems: Starmie~\cite{starmie} and SANTOS~\cite{santos}. Starmie discovers unionable tables via column embeddings from pre-trained language models. SANTOS uses open and synthesized knowledge bases to match column relationships within tables.   
We evaluate the effectiveness of these systems using three benchmarks: 
\texttt{$D^3L$ Small}\footnote{\url{https://github.com/alex-bogatu/DataSpiders}}~\cite{d3l}, 
\texttt{TUS Small}\footnote{\url{https://github.com/RJMillerLab/table-union-search-benchmark}}~\cite{Fatemeh18}, and 
\texttt{SANTOS Small}\footnote{\url{https://github.com/northeastern-datalab/santos}}~\cite{santos}. 
\texttt{$D^3L$ Small} is a collection of $654$ real-world tables, where each table in the data lake is manually annotated with all other related tables. \texttt{TUS Small} and \texttt{SANTOS Small} are collections of $1,539$ and $550$ synthetic tables, respectively, generated using random horizontal and vertical partitioning from real-world tables. The statistics of these benchmarks are shown in Table \ref{tab:data_discovery_dataset_stats}.
Note that the \texttt{$D^3L$ Small} is the most challenging because it was manually annotated rather than synthetically generated, has the highest number of rows per table on average, and has a high number of unionable tables of 110 on average. Furthermore, because SANTOS requires additional manual annotation of intent columns for all tables, we followed the authors' recommendation of using the first columns as intent columns for \texttt{$D^3L$ Small}.

\subsubsection{Accuracy of Data Discovery}
These benchmarks perform a data discovery query for table unionability for a query table $T$. The accuracy of the system in performing this query is measured in terms of Precision$@k$ and Recall$@k$ for different values of $k$, such that $k$ is the number of desired unionable tables to $T$~\cite{d3l, Fatemeh18, santos}. 
The precision and recall are calculated for each $k$ as the averages over $N$  query tables. 
In addition, we evaluated {\sysname}'s scalability using \texttt{SANTOS Large}~\cite{santos}, which contains $11,090$ tables without ground truth. 
In our evaluation, we used the same values of $(N,k)$ as in previous evaluations. More specifically, for \texttt{$D^3L$ Small} we used $(50, 185)$, for \texttt{TUS Small} we used $(150, 60)$, and for \texttt{SANTOS Small} we used $(50, 10)$.\shorten

\begin{table}[t]
\vspace*{-2ex}
\centering
\caption{Preprocessing and average query time for all benchmarks. {\sysname} outperforms Starmie and SANTOS in both.}
\vspace*{-2ex}
\label{tab:scalability_vs_santos}
\begin{tabular}{lllll} 
\toprule
\textbf{Benchmark}                                                                                  & \textbf{Time}       & \textbf{SANTOS} & \textbf{Starmie} & \textbf{KGLiDS}   \\ 
\midrule
\multirow{2}{*}{\begin{tabular}[l]{@{}l@{}}\textbf{$\mathbf{D^3L}$}\\\textbf{Small}\end{tabular}} & Preprocessing  & 2.96 hr         & 0.88 hr          & \textbf{0.37 hr}  \\
                                                                                                    & Avg. Query     & 18.8 s          & 0.71 s           & \textbf{0.02 s}   \\ 
\midrule
\multirow{2}{*}{\begin{tabular}[l]{@{}l@{}}\textbf{TUS}\\\textbf{Small}\end{tabular}}            & Preprocessing  & 4.66 hr         & 0.70 hr          & \textbf{0.69 hr}  \\
                                                                                                    & Avg. Query     & 13.35 s         & 0.04 s           & \textbf{0.02 s}   \\ 
\midrule
\multirow{2}{*}{\begin{tabular}[l]{@{}l@{}}\textbf{SANTOS}\\\textbf{Small}\end{tabular}}          & Preprocessing  & 1.89 hr         & 0.33 hr          & \textbf{0.26 hr}  \\
                                                                                                    & Avg. Query     & 17.12 s         & 0.11 s           & \textbf{0.01 s}   \\ 
\midrule
\multirow{2}{*}{\begin{tabular}[l]{@{}l@{}}\textbf{SANTOS}\\\textbf{Large}\end{tabular}}          & Preprocessing  & 30.40 hr        & 7.67 hr          & \textbf{4.15 hr}  \\
                                                                                                    & Avg. Query     & 12.28 s         & 0.79 s           & \textbf{0.24 s}   \\
\bottomrule
\end{tabular}
\vspace*{-2ex}
\end{table}

\begin{figure}
\vspace*{-2ex}
 \centering
\includegraphics[width=\columnwidth]{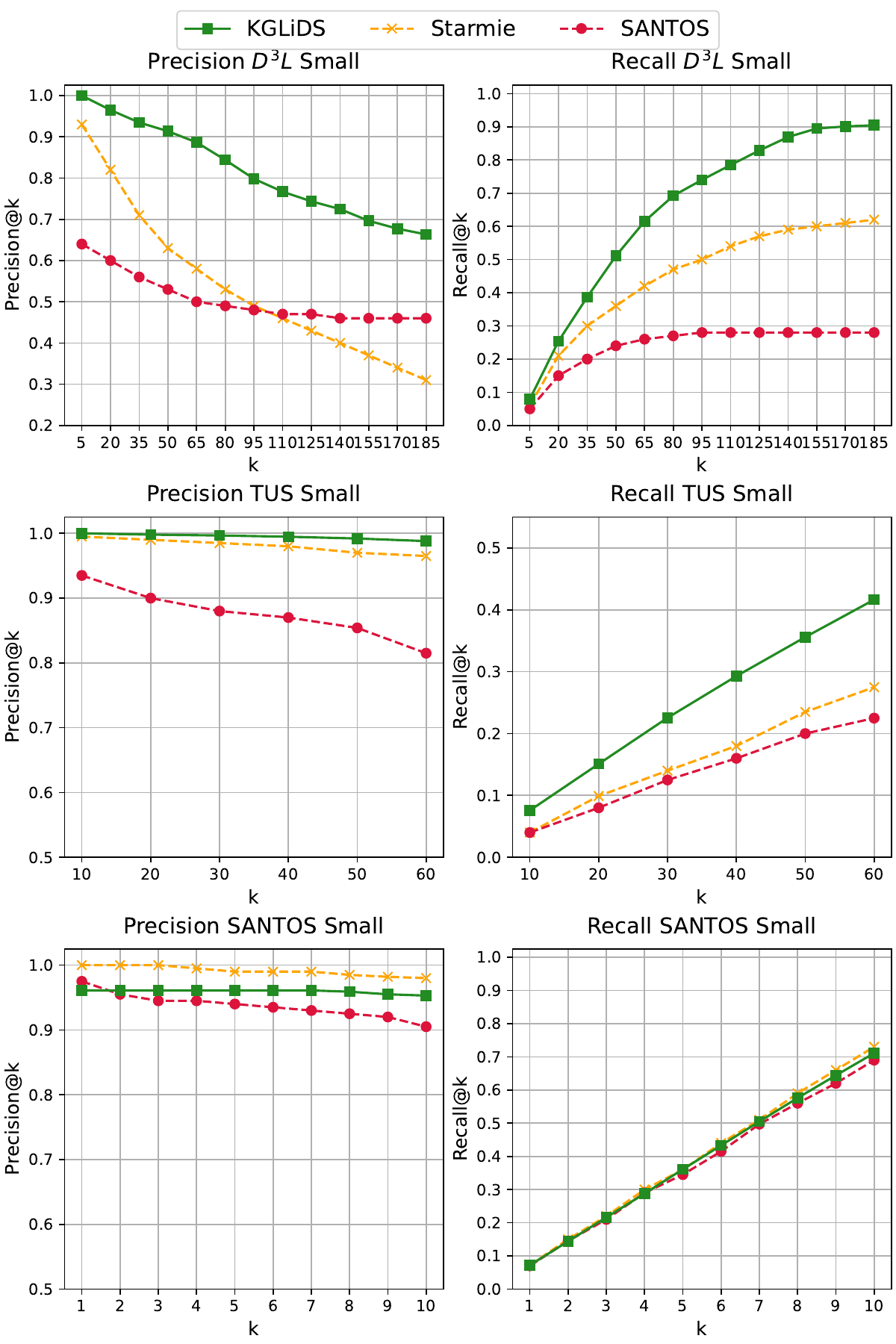} 
\vspace*{-4ex}
  \caption{Average precision and recall of unionable table discovery on all benchmark datasets.}
\label{fig:experiment_precision_recall}  
\vspace*{-4ex}
\end{figure}

Figure~\ref{fig:experiment_precision_recall} illustrates the average Precision@k and Recall@k across various systems and benchmarks. {\sysname} significantly outperforms Starmie and SANTOS in precision and recall on $D^3L$ \texttt{Small} and \texttt{TUS Small}. Unlike {\sysname}, Starmie uses column embeddings from pre-trained language models, which work well with textual columns but do not achieve the same accuracy for numerical columns. We analyzed Starmie's individual column matches for $D^3L$ \texttt{Small}, which had 52.2 precision for numerical columns compared to 63.4 for textual columns. Our CoLR models are trained to predict column similarity even with different value distributions, which enhances similarity prediction for real-world benchmarks like $D^3L$ \texttt{Small}. 
{\sysname} still achieves high accuracy for synthetically generated benchmarks, where unionable tables have the same distribution. 
The \texttt{Santos Small} benchmark used tables of relatively smaller size and a low value of $k$. Hence, the three systems archive comparable precision and recall.

\subsubsection{System Performance Analysis} 
These experiments analyze the preprocessing overhead and execution time of the query tables. The preprocessing overhead is the offline time required to profile tables in a data lake. The query execution time is reported as the average query time of all query tables (see Table \ref{tab:data_discovery_dataset_stats} for the number of query tables in each benchmark). 
Table~\ref{tab:scalability_vs_santos} summarizes our results. {\sysname} consistently outperforms SANTOS and Starmie in preprocessing and average query times for small and large benchmarks. We focus on the \texttt{SANTOS Large} benchmark as it has the largest number of tables.
{\sysname} is 7.3x faster than SANTOS in preprocessing time. This is because SANTOS analyzes data lake tables on the granularity of column values, matching each value against two knowledge bases (KB): an open KB (YAGO \cite{yago}) and a synthesized KB generated during preprocessing. SANTOS then iterates over \textit{all value pairs} of matching columns per table to determine their semantic relationships.\shorten 

{\sysname} analyzes data lakes on the granularity of columns and applies our CoLR models to column samples. 
Compared to Starmie, {\sysname} achieves a 1.8x speed up in preprocessing time. 
The preprocessing phase of Starmie includes training a language model (LM) to generate the column embeddings for a data lake. Starmie uses data augmentation on data lake columns to generate enough samples to train the LM. Further, Starmie needs to train the LM for a sufficient number of epochs to achieve a decent score (we use ten epochs as recommended by the authors of Starmie). Unlike Starmie, {\sysname} does not require training our embedding models per data lake as our models are independently pre-trained on open datasets.\shorten 

Furthermore, {\sysname} converts data discovery queries into SPARQL queries against our LiDS graph. We designed these queries to leverage the built-in indices in RDF engines. Hence, {\sysname} is 51.2x faster than SANTOS in the \texttt{SANTOS Large} benchmark, as shown in Table \ref{tab:scalability_vs_santos}. For a table unionability query, SANTOS retrieves a list of candidate unionable tables by looking up two indices (one for each KB) for column relationships similar to those found between pairs of columns in the query table. Then, the query table is matched against candidate tables at different granularities to determine the unionability score for each candidate. This results in many comparisons and higher query times for SANTOS. Similarly, {\sysname} is 3.3x faster than Starmie in average query time. While Starmie uses an efficient vector index (Hierarchical Navigable Small World (HNSW) \cite{hnsw}) to retrieve similar columns based on embeddings, it requires calculating an approximate distance between embeddings of size 768.

\begin{figure}[t]
\vspace*{-2ex}
\centering
\includegraphics[width=0.9\columnwidth]{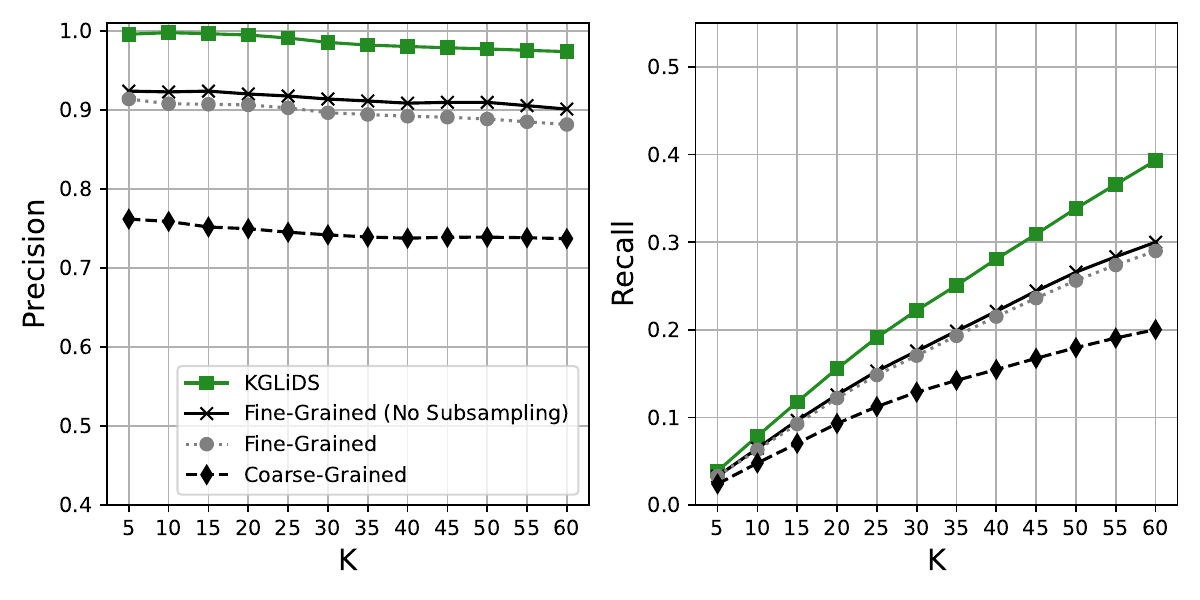} 
\vspace*{-2ex}
  \caption{Ablation study for the effectiveness of {\sysname} for table union search on the \texttt{TUS Small} benchmark.}
  \label{fig:ablation_study}
\end{figure}

\subsubsection{Ablation Study}
We conducted an ablation study to analyze our models for generating column and label embeddings. Figure \ref{fig:ablation_study} shows the precision and recall of different KGLiDS configurations on the \texttt{TUS Small} benchmark. The highest scores were achieved by utilizing both the CoLR and label embeddings (indicated by the green line). In existing benchmarks, column names are a strong predictor of table unionability. However, raw value-based unionability is useful for data lakes without column names (e.g. anonymized). Figure~\ref{fig:ablation_study} illustrates that KGLiDS still performs competitively when using only CoLR embeddings w.r.t Starmie and SANTOS, whose TUS results shown in Figure~\ref{fig:experiment_precision_recall}.
This predicts unionability based only on raw values (indicated as ``Fine-Grained") in Figure~\ref{fig:ablation_study}.  

We developed coarse-grained embedding models inspired by \cite{MuellerS19}, which introduced an embedding-based method through three coarse-grained models. We evaluated these coarse-grained embedding models and our fine-grained CoLR models. The results indicate that our fine-grained CoLR models significantly improved precision and recall, with precision reaching up to 17\% higher and recall up to 50\% higher at $k=60$, as illustrated in Figure~\ref{fig:ablation_study}.
We analyzed the impact of subsampling columns when generating column embeddings using our CoLR models, as outlined in Section~\ref{sec:data_profiling}. Using the CoLR models on column samples representing 10\% of the original column sizes demonstrates comparable precision and recall to using them on entire columns. Our subsampling approach reduces the profiling time for the TUS Small benchmark by 14\%. 

\subsection{{\sysname} and Pipeline Abstraction}
\label{sec:evaluation-pipelineGraphs}

We evaluate \sysname{} against GraphGen4Code~\cite{graph4code}, which is a toolkit to generate a knowledge graph for code~\footnote{GrapGen4Code is obtained from \url{https://github.com/wala/graph4code}}. 
We collected $13,800$ data science pipeline scripts used in the top $1000$ datasets from Kaggle, which includes $3,775$ tables and $141,704$ columns \footnote{The script to download these datasets is available at {\sysname}' repository}. 
We selected these pipelines and datasets based on the number of user votes on the Kaggle platform. For each dataset, we select up to 20 most voted Python pipelines. The datasets are related to various supervised and unsupervised tasks in different domains, such as health, economics, games, and product reviews.
Table \ref{tab:graph_analysis} summarizes the graph sizes and processing time for generating the {\graphname} and GraphGen4Code graphs for $13,800$ pipelines. 
GraphGen4Code is developed for general semantic code abstraction. In {\sysname}, we capture semantics related to data science artifacts only using lightweight static code augmented by our documentation and dataset usage analyses. Hence, {\sysname} achieves a graph reduction of more than 82\% in 95\% less time for the same set of pipelines.

Table~\ref{tab:modelled_aspects} illustrates a breakdown of the graphs generated by both systems. GraphGen4Code is a general-purpose tool for code abstraction. Hence, it captures irrelevant information to data science artifacts. For example, approximately 30\% of the GraphGen4Code graph includes local syntactical script information, namely, \emph{statement location} and \emph{function parameter order}. In contrast, {\sysname} captures semantics related primarily to data science, including ones not modelled by GraphGen4Code, such as \emph{dataset reads} that indicate when \texttt{CSV} files are read, or \emph{library hierarchy} that model the structure of functions and classes in a data science library. These aspects are essential for automating data science pipelines, such as predicting a near-optimal operation for data cleaning or transformation. In addition, all nodes in the {\graphname} graph are labeled with an RDF type, a standard requirement when constructing knowledge graphs for facilitating graph queries, building a graph schema, and enabling reasoning by RDF engines.\shorten

\begin{table}[t]
\vspace*{-2ex}
\centering
\caption{Comparison of the RDF graphs and analysis time for {\sysname} and GraphGen4Code on a collection of 13,800 data science pipelines from Kaggle.}
\vspace*{-2ex}
\label{tab:graph_analysis}
\begin{tabular}{lll}
\toprule
Statistic                    & KGLiDS & GraphGen4Code \\
\midrule
No. triples (edges)         & 16,640,400 & 97,537,947    \\
No. unique nodes     & 3,053,416  & 20,737,622   \\
No. unique edges     & 16    & 16           \\
Size            & 1.49 GB   & 16.55 GB          \\
Analysis time & 1.9 hr & 37.59 hr  \\
\bottomrule
\end{tabular}
\vspace*{-3ex}
\end{table}



\subsection{Evaluating Our On-Demand Automation}
\label{sec:evaluationDataPrep}
This section presents our evaluation for \sysname{} against SOTA systems in data cleaning, transformation and AutoML. For systems evaluation, we used a total of {\totalDatasets} \textbf{unseen} datasets collected from different benchmarks. A complete list of the {\totalDatasets} datasets, including their ID and statistics, is available at our {\sysname} repository\footnote{\url{https://github.com/CoDS-GCS/kglids/tree/master/gnn_applications}}. Each dataset is associated with an ML task. For example, we perform data cleaning using one of the systems and then model the task associated with the dataset. In this case, we clean the null values in the age column in the Titanic dataset and then train a classification model to predict survival.  We consider the accuracy of the trained model as an indicator of the accuracy of each system, i.e., the better the cleaning, the higher the accuracy will be.\shorten

\subsubsection{Data Cleaning} 
This experiment analyzes our on-demand cleaning against \holoclean{}, the SOTA general cleaning platform. \holoclean{} uses statistical learning and inference to unify a range of data-repairing methods. We used the most recent version of \holoclean{}, a.k.a Aimnet~\cite{aimnet}, where a user does not need to specify a set of denial constraints. We used it's \textit{null detector} only with its default settings in \holoclean{}'s GitHub repository~\footnote{\url{https://github.com/HoloClean/holoclean/tree/latest-aimnet}}.
We used the 13 datasets from an AutoML benchmark~\cite{kgpip} that contained missing values and supplemented them with five datasets from the UCI repository~\cite{uci}. Our evaluation consisted of cleaning these datasets using \sysname{} and \holoclean{} and then training a random forest classifier. We also consider a baseline approach that performs modelling by dropping null values. The metric we use to evaluate on-demand data cleaning is the F1 scores of the random forest classifier trained using cross-validation over ten folds on the cleaned datasets. We also evaluate the systems in terms of execution time and memory usage.\shorten

\begin{table}[t]
\vspace*{-2ex}
\caption{A breakdown of the graphs generated by {\sysname} and GraphGen4Code. The number and percentage of triples modeling each aspect are shown.}
\vspace*{-2ex}
\label{tab:modelled_aspects}
\begin{tabular}{lll|ll}
\toprule
Modelled Aspect      & \multicolumn{2}{c}{\sysname} & \multicolumn{2}{c}{GraphGen4Code}                                             \\
\midrule
Dataset reads               & 0.03M      & 00.2\%    &  -        & -         \\
Library hierarchy           & 0.02M      & 00.1\%    &  -        & -         \\
RDF node types              & 2.55M   & 15.3\%    &   -          & -           \\
Statement location          &  -        & -         & 3.97M      & 04.1\%    \\
Variable names              &  -        & -         & 0.99M      & 01.0\%     \\
Func. parameter order    &   -       & -         & 25.13M        & 25.8\%     \\
Column reads                & 3.52M   & 21.1\%    & 2.00M        & 02.0\%   \\ 
Library calls               & 0.51M     & 03.0\%    & 15.27M     & 15.6\%   \\
Code flow                   & 2.11M   & 12.7\%    & 20.30M       & 20.8\%   \\
Data flow                   & 1.27M   & 07.6\%    & 13.30M       & 13.6\%   \\
Control flow type           & 0.81M     & 04.9\%    & 1.12M      & 01.2\%     \\
Func. parameters         & 3.72M   & 22.4\%    & 7.51M           & 07.7\%   \\
Statement text              & 2.12M   & 12.7\%    & 7.94M        & 08.1\%    \\
\midrule
Total                       & 16.64M  & 100\%          & 97.54M    & 100\% \\     

\bottomrule
\end{tabular}
\vspace*{-3ex}
\end{table}

\begin{figure}[t]
\vspace{-2ex}
  \centering
    \includegraphics[width=\columnwidth]{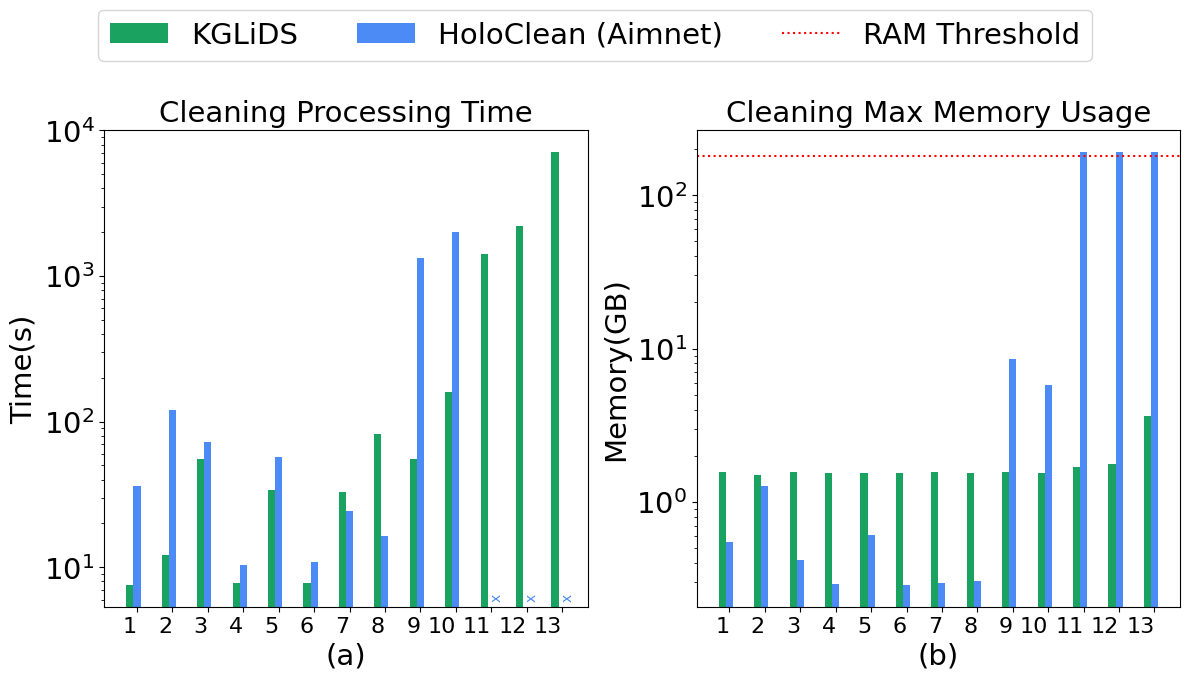}
    \vspace*{-4ex}
    \caption{The performance of {\sysname} vs \holoclean{} (Aimnet) on the 13 datasets using a VM with 189 GB RAM. The X-axis represents the dataset ID, and the Y-axis is the time (a) or memory usage (b) consumed by each system. Datasets are sorted by size in increasing order. {\sysname} reduces the time significantly while using an almost fixed amount of memory across datasets, e.g., less than 4GB RAM.\shorten 
    } 
  \label{fig:exp_data_cleaning}
  \vspace{-2ex}
\end{figure}

Table~\ref{tab:cleaning_results} shows that \sysname{} achieves consistently comparable or better F1 scores than \holoclean{}. However, \sysname{} outperforms significantly \holoclean{} in $85\%$ of datasets in execution time and $38\%$ datasets in memory usage, as illustrated in Figure \ref{fig:exp_data_cleaning}. \holoclean{} failed with out-of-memory errors while attempting to clean datasets $\#11$, $\#12$, and $\#13$, and could not complete the cleaning process for these datasets. \holoclean{} generates multiple tables containing dataset information throughout its cleaning process. Therefore, its memory requirements increase as the dataset size increases. In contrast, \sysname{}'s memory usage does not increase significantly as our models use fixed-size embeddings regardless of the table size. This highlights a key advantage of our on-demand cleaning approach, which enhances modeling performance while training specific tasks on a dataset. 
This method avoids excessive computations required by \holoclean{}. 

\subsubsection{Data Transformation} 

We compare \sysname{} against \autolearn{} \cite{autolearn}, a regression-based feature learning algorithm to predict data transformation.
In this evaluation, we used 17 datasets from~\autolearn{}'s experimental evaluation~\cite{autolearn}, which is available in the UCI repository~\cite{uci}. These datasets were assigned IDs from $14$ to $30$. 
Similar to data cleaning, our evaluation process includes applying the data transformation recommended by \sysname{} and \autolearn{} to each dataset, and then we trained a random forest classifier.
The \autolearn{} paper reported the accuracy scores. Hence, we use it as the evaluation metric for the random forest classifier trained using cross-validation over 5-fold on the transformed datasets. We also evaluate the time and memory usage of these systems. We did not manage to reproduce the reported results of \autolearn{} using the default parameters of its GitHub repository~\footnote{\url{https://github.com/saket-maheshwary/AutoLearn/tree/master}}. We tuned hyperparameters for reported accuracy within a three-hour time limit. \sysname{} consistently matches or surpasses \autolearn{} accuracy, as shown in Table~\ref{tab:transformation_results}.\shorten 

\begin{table}[t]

\vspace*{-2ex}
 
  \centering
  \caption{F1-Scores for Data Cleaning. The performance of {\sysname} vs \holoclean{} (Aimnet) using multiple ML tasks on 13 datasets. {\sysname} slightly outperforms \holoclean{} in small datasets while \holoclean{} encountered an out-of-memory ($OOM$) issue when processing large datasets.}
  \vspace*{-1ex}
  \label{tab:cleaning_results}
  \begin{adjustbox}{width=0.5\textwidth}
    \begin{tabular}{llll}
    \toprule
    ID - Dataset & Baseline & Holoclean & KGLiDS \\
    \midrule
    1 - hepatitis & \textbf{69.76} & 67.78 & 69.35 \\
    2 - horsecolic & 00.00 & 82.28 & \textbf{85.38} \\
    3 - housevotes84 & 96.10 & \textbf{96.64} & 95.89 \\
    4 - breastcancerwisconsin & \textbf{97.43} & 95.93 & 96.85 \\
    5 - credit & 88.11 & 86.95 & \textbf{88.17} \\
    6 - cleveland\_heart\_disease & \textbf{28.31} & 27.51 & 25.50 \\
    7 - titanic & 70.68 & 81.89 & \textbf{82.63} \\
    8 - creditg & 00.00 & 65.63 & \textbf{66.63} \\
    9 - jm1 & \textbf{61.59} & 60.55 & 61.55 \\
    10 - adult & 79.15 & 78.49 & \textbf{79.46} \\
    11 - higgs & 71.70 & $OOM$ & \textbf{71.73} \\
    12 - APSFailure & \textbf{91.49} & $OOM$ & 90.89 \\
    13 - albert & 00.00 & $OOM$ & \textbf{66.70} \\
    \bottomrule
  \end{tabular}
  \end{adjustbox}
\end{table}

\begin{table}[t]
\vspace*{-2ex}
\caption{Accuracy for Data Transformation. The performance of {\sysname} vs AutoLearn on 17 datasets associated with multiple ML classification tasks. Autolearn results are formatted as Y(X) where Y is the reported accuracy in \cite{autolearn} and X is the outcome of reproducing Autolearn experiments, $TO$ if Autolearn times out in three hours, or out-of-memory ($OOM$).\shorten }
\vspace*{-1ex}
\begin{adjustbox}{width=0.48\textwidth}
\label{tab:transformation_results}
\begin{tabular}{llllllll}

\toprule
ID - Dataset & Baseline & Autolearn & KGLiDS \\
\midrule
14 - fertility\_Diagnosis&82.00&84.00 (\textbf{86.12})&85.00\\
15 - haberman&68.63&65.34 (71.89)&\textbf{71.92}\\
16 - wine&96.07&97.20 (\textbf{98.33})&97.17\\
17 - Ecoli&82.73&86.59 (81.23)&\textbf{88.10}\\
18 - pima diabetes&\textbf{75.37}&73.05 (75.13)&75.14\\
19 - Banke Note&99.05&99.56 (\textbf{99.93})&98.91\\
20 - ionosphere&93.15&92.30 (\textbf{93.46})&93.44\\
21 - sonar&73.55&77.87 (78.83)&\textbf{78.86}\\
22 - Abalone&22.91&22.21 (\textbf{24.96})&24.56\\
23 - libras&71.94&70.22 (79.13)&\textbf{81.39}\\
24 - waveform&82.10&81.12 ($TO$)&\textbf{85.00}\\
25 - letter recognition&93.96&94.14 ($TO$)&\textbf{96.46}\\
26 - opticaldigits&96.38&96.57 ($TO$)&\textbf{98.10}\\
27 - featurepixel&95.5&94.20 ($TO$)&\textbf{97.65}\\
28 - shuttle&\textbf{99.97}&99.81 ($TO$)&99.96\\
29 - featurefourier&79.9&79.31 ($TO$)&\textbf{82.55}\\
30 - poker &68.1&72.26 ($OOM$)&\textbf{75.32}
\\
\bottomrule
\end{tabular}
\end{adjustbox}
\vspace*{-3ex}
\end{table}

\begin{figure}[t]
\vspace{-1ex}
\centering
\includegraphics[width=\columnwidth]{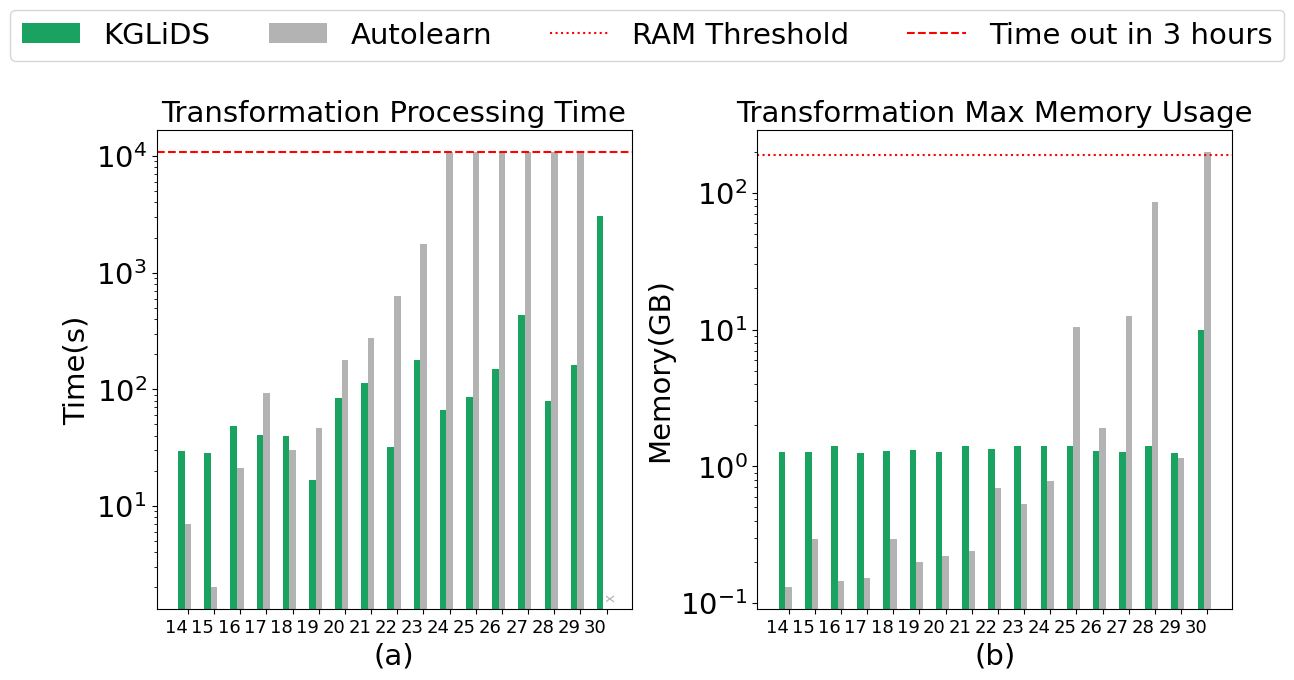} 
\vspace*{-4ex}
  \caption{
  The performance of {\sysname} vs \autolearn{} on the 17 datasets using a VM with 189 GB RAM. The X-axis represents the dataset ID, and the Y-axis is the time (a) or memory usage (b) consumed by each system. Datasets ascend by size. {\sysname} significantly outperforms \autolearn{} in time with nearly constant memory usage of less than 10 GB.\shorten
  }
  \label{fig:transformation_autolearn}
  \vspace*{-4ex}
\end{figure}

\sysname{} significantly outperformed \autolearn{} in execution time with stable memory usage as data size grows, as shown in Figure~\ref{fig:transformation_autolearn}. \autolearn{} employs distance correlation to identify pairwise correlated features, classify them into linear and non-linear correlations, and then generate informative new features. 
The original dataset's row and feature count, inter-feature correlations, created features, and the chosen feature quantity collectively impact \autolearn{}'s memory usage. Hence, the dataset's absolute size is not the primary factor influencing memory usage in \autolearn{}'s transformation. Unlike \autolearn{}, \sysname{} uses fixed-size embeddings, making predictions independent of table size.\shorten

\begin{figure}[t]
 \centering
  \includegraphics[width=\columnwidth]{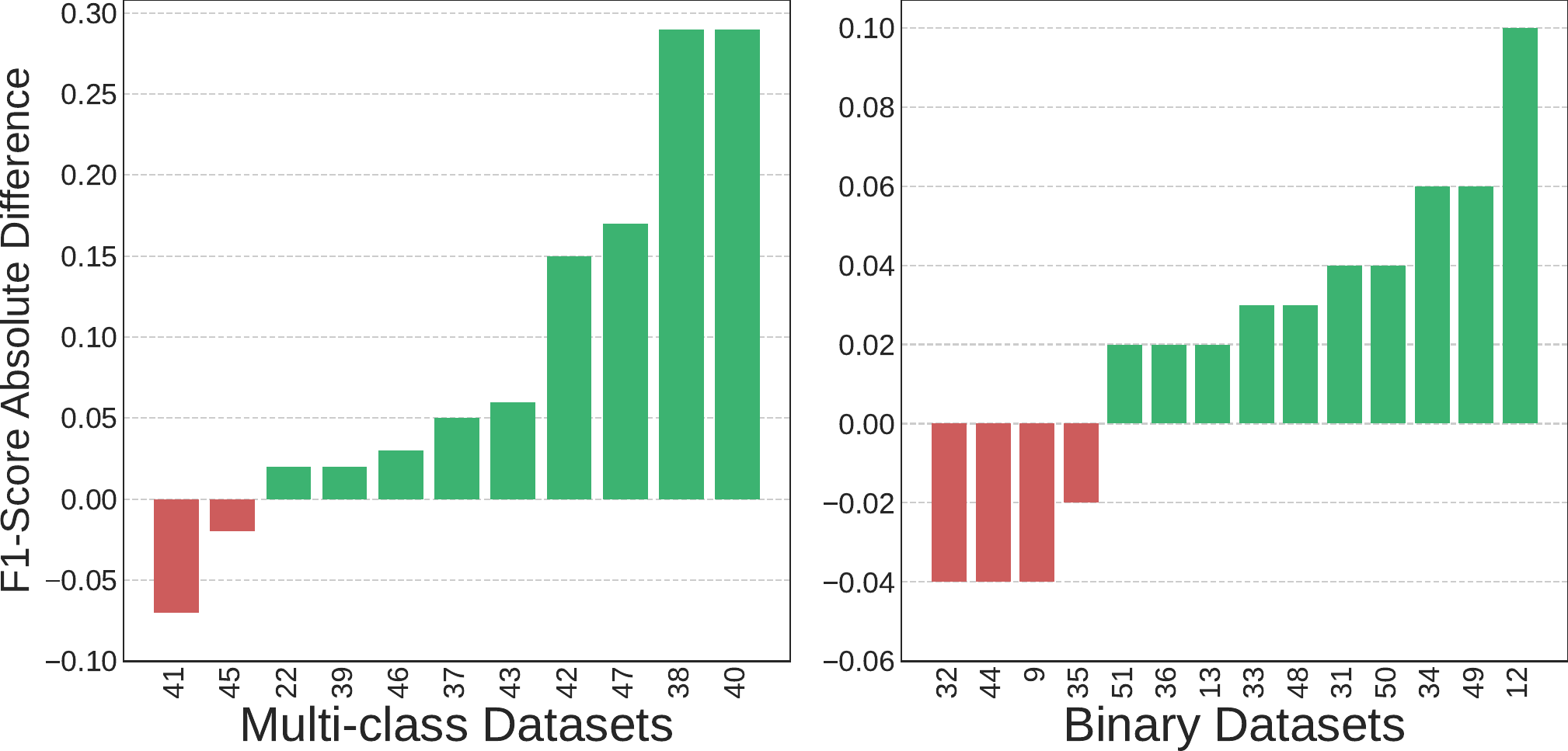} 
  \caption{ The F1-Score difference between KGpip with KGLiDS ($Pip_{LiDS}$) and KGpip with GraphGen4Code ($Pip_{G4C}$) across 24 AutoML datasets. The X-axis denotes the dataset IDs. 
  $Pip_{LiDS}$ efficiently prunes hyperparameter search space and helps KGpip improve F1-Scores in most cases.
  }
  \label{fig:automl_experiment}
  \vspace*{-3ex}
\end{figure}

\subsubsection{AutoML and Better Hyperparameter Search}
We integrated the revised version of the KGpip pipeline into {\sysname}. The inference pipeline utilizes {\graphname} to recommend a starting point for the hyperparameter search. The recommended initial hyperparameters by {\sysname} for a given dataset are the most commonly used for the top-voted pipelines associated with the most similar dataset found in the {\graphname} graph. 
We used 24 benchmark tables used by the KGpip \cite{kgpip} system. The datasets are real-world datasets from a diversity of sources and problem domains. These datasets are associated with different binary and multi-class classification tasks of enough difficulty for the learning algorithms.\shorten 

Figure \ref{fig:automl_experiment} shows a comparison in the F-Score of the original KGpip pipeline ($Pip_{G4C}$) using GraphGen4Code and our revised pipeline ($Pip_{LiDS}$) using our {\graphname} graph. We follow the same evaluation procedure in~\cite{kgpip} and limit the time budget to 40 seconds to avoid the exploration of the full search space. 
$Pip_{LiDS}$ outperforms $Pip_{G4C}$ in the majority of datasets. In addition, a \textit{two-tailed t-Test} between the scores obtained from both pipelines shows that $Pip_{LiDS}$ has significantly better scores with a t-Test value of $0.012$ ($p < 0.05$). Using {\graphname} enables AutoML systems to leverage data scientists' accumulated knowledge to enhance scores and reduce time for hyperparameter exploration.\shorten



\section{Related Work}
\label{sec:related_word}

{\sysname} represents our pioneering effort in developing the data science knowledge graph proposed in~\cite{KEK} to interlink data science across federated datasets.

\nsstitle{Data Discovery Systems} profile datasets to construct different navigational data structures. Starmie \cite{starmie} uses embeddings from language models with an HNSW \cite{hnsw} index. SANTOS \cite{santos} builds column relationship trees by matching column values against open and synthetic knowledge bases.
$D^3L$~\cite{d3l} constructs a variety of feature vectors for each column and matches them using hash-based indices.
These systems rarely use open standards. We are the first to design navigational data structures capturing the semantics of datasets and pipelines based on knowledge graph technologies. Furthermore, we enable several data discovery operations, such as unionable tables, joinable tables, and join path discovery. Unlike \cite{ZhangI20, ZhuDNM19, Fatemeh18}, {\sysname} also enables users to query datasets based on embeddings and combined easily different operations in one pipeline script. Our KGLac~\cite{KGLacDemo} demonstrated the capabilities of {\sysname}'s data discovery support.\shorten

\nsstitle{Table representation learning (TRL)} aims to construct fixed-size, low-dimensional representations for tables. Traditional approaches construct table or column embeddings using feature-based ML models \cite{sato} or deep neural networks \cite{MuellerS19}. The approach of \cite{MuellerS19} was originally proposed for variable recognition. {\sysname} adapts this approach and improves it by training fine-grained CoLR models and using them to predict relationships between data items. More recent approaches such as \cite{embeddings_of_rel_datasets, turl, tabtransformer, tabbie} generate TRL using language models. These approaches are orthogonal to {\sysname}, which is modular for integrating new models seamlessly.

\nsstitle{Capture Pipeline Semantics.} 
Several approaches~\cite{scro,code_ontology,hmm_java_ontology,graph4code} have been proposed to provide a semantic representation of source code with knowledge graphs. 
These techniques cannot be generalized to data science pipelines due to their heavy reliance on the detailed static analysis of Java, where information such as method input and return types is straightforward to determine. 
GraphGen4Code \cite{graph4code} is a toolkit utilizing a tool called WALA~\cite{WALA} for general-purpose static code analysis.
Unlike GraphGen4Code, {\sysname} combines static code analysis with library documentation and dataset usage analyses to have a rich semantic abstraction that captures the essential concepts in data science pipelines.\shorten

\nsstitle{Learning from data science pipelines} has been utilized by different systems. 
Auto-Suggest~\cite{Auto-Suggest} learns from data science notebooks to recommend Python scripts for database operations, such as Join, Pivot, Unpivot, and GroupBy. 
Auto-Pipeline~\cite{autopipeline} uses a by-target approach to automate table manipulation operations, such as Join and Group-by.
ModsNet~\cite{topk} aims to select the top-K pre-trained models from a set of data science models given an example dataset. ModsNet is based on GNN~\cite{gao2023survey} and is not optimized to automate data science pipelines in terms of data cleaning and transformation. 
Unlike all the above systems, {\sysname} captures the semantics of data science pipelines and offers holistic support for data discovery and on-demand pipeline automation.






\section{Conclusion}
\label{sec:conclusion}

This paper proposed a scalable platform (KGLiDS) that employs machine learning to abstract the semantics of data science artifacts and construct a knowledge graph (KG) interconnecting them. {\sysname} developed an advanced data profiler empowered by machine learning to analyze data items, including datasets, tables, and columns. {\sysname} further implements specialized static code analysis that infers information not otherwise obtainable with general-purpose static analysis, resulting in a richer, more accurate, and more compact abstraction of data science pipelines. 
{\sysname} enables novel use cases for discovery, exploration, reuse, and automation in data science platforms. 
Our comprehensive evaluation uses 4 data discovery benchmarks and 51 unseen datasets collected from different AutoML benchmarks to compare KGLiDS against the SOTA systems in data discovery, data cleaning, transformation, and AutoML. Our experiments show that KGLiDS consumes significantly less time w.r.t the SOTA systems while achieving comparable or better accuracy. 
In our in-progress and future work, we are incorporating LLMs into our {\sysname} system for more use cases, such as exploratory data analysis, feature selection and engineering.

\nsstitle{Acknowledgement.}
We thank Dr. Renee Miller and her team for their guidance in replicating their results and selecting optimal configurations for the Starmie and SANTOS systems, especially for the $D^3L$ benchmark. 
We also thank Dr. Ihab Ilyas for his advice on which version of HoloClean to use.

\clearpage

\balance
\bibliographystyle{ACM-Reference-Format}
\bibliography{references}


\begin{thebibliography}{63}


\ifx \showCODEN    \undefined \def \showCODEN     #1{\unskip}     \fi
\ifx \showDOI      \undefined \def \showDOI       #1{#1}\fi
\ifx \showISBNx    \undefined \def \showISBNx     #1{\unskip}     \fi
\ifx \showISBNxiii \undefined \def \showISBNxiii  #1{\unskip}     \fi
\ifx \showISSN     \undefined \def \showISSN      #1{\unskip}     \fi
\ifx \showLCCN     \undefined \def \showLCCN      #1{\unskip}     \fi
\ifx \shownote     \undefined \def \shownote      #1{#1}          \fi
\ifx \showarticletitle \undefined \def \showarticletitle #1{#1}   \fi
\ifx \showURL      \undefined \def \showURL       {\relax}        \fi
\providecommand\bibfield[2]{#2}
\providecommand\bibinfo[2]{#2}
\providecommand\natexlab[1]{#1}
\providecommand\showeprint[2][]{arXiv:#2}

\bibitem[\protect\citeauthoryear{??}{WAL}{2022}]%
        {WALA}
 \bibinfo{year}{2022}\natexlab{}.
\newblock \bibinfo{title}{WALA Tool. Accessed: 2022-07-15}.
\newblock
\newblock
\urldef\tempurl%
\url{https://wala.github.io}
\showURL{%
\tempurl}


\bibitem[\protect\citeauthoryear{??}{gra}{2023}]%
        {graphdb}
 \bibinfo{year}{2023}\natexlab{}.
\newblock \bibinfo{title}{The GraphDB RDF Engine. Accessed: 2023-12-01}.
\newblock
\newblock
\urldef\tempurl%
\url{https://www.ontotext.com/products/graphdb/}
\showURL{%
\tempurl}


\bibitem[\protect\citeauthoryear{??}{kag}{2023}]%
        {kaggle}
 \bibinfo{year}{2023}\natexlab{}.
\newblock \bibinfo{title}{Kaggle Portal. Accessed: 2023-12-01}.
\newblock
\newblock
\urldef\tempurl%
\url{https://www.kaggle.com/}
\showURL{%
\tempurl}


\bibitem[\protect\citeauthoryear{??}{kag}{2024}]%
        {kaggle_survey}
 \bibinfo{year}{2024}\natexlab{}.
\newblock \bibinfo{title}{Kaggle ML Survey 2022. Accessed: 2024-02-01}.
\newblock
\newblock
\urldef\tempurl%
\url{https://www.kaggle.com/kaggle-survey-2022}
\showURL{%
\tempurl}


\bibitem[\protect\citeauthoryear{Abdelaziz, Dolby, McCusker, and
  Srinivas}{Abdelaziz et~al\mbox{.}}{2021}]%
        {graph4code}
\bibfield{author}{\bibinfo{person}{Ibrahim Abdelaziz}, \bibinfo{person}{Julian
  Dolby}, \bibinfo{person}{James McCusker}, {and} \bibinfo{person}{Kavitha
  Srinivas}.} \bibinfo{year}{2021}\natexlab{}.
\newblock \showarticletitle{A Toolkit for Generating Code Knowledge Graphs}.
\newblock \bibinfo{journal}{\emph{Proceedings of Knowledge Capture Conference
  (K-CAP)}} (\bibinfo{year}{2021}).
\newblock
\urldef\tempurl%
\url{https://doi.org/10.1145/3460210.3493578}
\showURL{%
\tempurl}


\bibitem[\protect\citeauthoryear{Abdelaziz, Mansour, Ouzzani, Aboulnaga, and
  Kalnis}{Abdelaziz et~al\mbox{.}}{2017}]%
        {lusail}
\bibfield{author}{\bibinfo{person}{Ibrahim Abdelaziz}, \bibinfo{person}{Essam
  Mansour}, \bibinfo{person}{Mourad Ouzzani}, \bibinfo{person}{Ashraf
  Aboulnaga}, {and} \bibinfo{person}{Panos Kalnis}.}
  \bibinfo{year}{2017}\natexlab{}.
\newblock \showarticletitle{Lusail: {A} System for Querying Linked Data at
  Scale}.
\newblock \bibinfo{journal}{\emph{Proceedings of the {VLDB} Endowment,
  (PVLDB)}} (\bibinfo{year}{2017}), \bibinfo{pages}{485--498}.
\newblock
\urldef\tempurl%
\url{http://www.vldb.org/pvldb/vol11/p485-abdelaziz.pdf}
\showURL{%
\tempurl}


\bibitem[\protect\citeauthoryear{Alnusair and Zhao}{Alnusair and Zhao}{2010}]%
        {scro}
\bibfield{author}{\bibinfo{person}{Awny Alnusair} {and} \bibinfo{person}{Tian
  Zhao}.} \bibinfo{year}{2010}\natexlab{}.
\newblock \showarticletitle{Component search and reuse: An ontology-based
  approach}. In \bibinfo{booktitle}{\emph{Proceedings of IEEE International
  Conference on Information Reuse \& Integration, (IRI)}}.
  \bibinfo{pages}{258--261}.
\newblock
\urldef\tempurl%
\url{https://doi.org/10.1109/IRI.2010.5558931}
\showURL{%
\tempurl}


\bibitem[\protect\citeauthoryear{Atzeni and Atzori}{Atzeni and Atzori}{2017}]%
        {code_ontology}
\bibfield{author}{\bibinfo{person}{Mattia Atzeni} {and}
  \bibinfo{person}{Maurizio Atzori}.} \bibinfo{year}{2017}\natexlab{}.
\newblock \showarticletitle{CodeOntology: RDF-ization of Source Code}. In
  \bibinfo{booktitle}{\emph{Proceedings of International Semantic Web
  Conference, (ISWC)}}. \bibinfo{pages}{20--28}.
\newblock
\showISBNx{978-3-319-68203-7}
\urldef\tempurl%
\url{https://doi.org/10.1007/978-3-319-68204-4_2}
\showURL{%
\tempurl}


\bibitem[\protect\citeauthoryear{Biessmann, Rukat, Schmidt, Naidu, Schelter,
  Taptunov, Lange, and Salinas}{Biessmann et~al\mbox{.}}{2019}]%
        {DataWig}
\bibfield{author}{\bibinfo{person}{Felix Biessmann}, \bibinfo{person}{Tammo
  Rukat}, \bibinfo{person}{Phillipp Schmidt}, \bibinfo{person}{Prathik Naidu},
  \bibinfo{person}{Sebastian Schelter}, \bibinfo{person}{Andrey Taptunov},
  \bibinfo{person}{Dustin Lange}, {and} \bibinfo{person}{David Salinas}.}
  \bibinfo{year}{2019}\natexlab{}.
\newblock \showarticletitle{DataWig: Missing Value Imputation for Tables}.
\newblock \bibinfo{journal}{\emph{Journal of Machine Learning Research}}
  \bibinfo{volume}{20}, \bibinfo{number}{175} (\bibinfo{year}{2019}),
  \bibinfo{pages}{1--6}.
\newblock
\urldef\tempurl%
\url{http://jmlr.org/papers/v20/18-753.html}
\showURL{%
\tempurl}


\bibitem[\protect\citeauthoryear{Bogatu, Fernandes, Paton, and
  Konstantinou}{Bogatu et~al\mbox{.}}{2020}]%
        {d3l}
\bibfield{author}{\bibinfo{person}{Alex Bogatu}, \bibinfo{person}{Alvaro
  Fernandes}, \bibinfo{person}{Norman Paton}, {and} \bibinfo{person}{Nikolaos
  Konstantinou}.} \bibinfo{year}{2020}\natexlab{}.
\newblock \showarticletitle{Dataset Discovery in Data Lakes}. In
  \bibinfo{booktitle}{\emph{Proceedings of International Conference on Data
  Engineering (ICDE)}}. \bibinfo{pages}{709--720}.
\newblock
\urldef\tempurl%
\url{https://doi.org/10.1109/ICDE48307.2020.00067}
\showURL{%
\tempurl}


\bibitem[\protect\citeauthoryear{Bursztyn, Goasdou{\'{e}}, Manolescu, and
  Roatis}{Bursztyn et~al\mbox{.}}{2015}]%
        {BursztynGMR15}
\bibfield{author}{\bibinfo{person}{Damian Bursztyn},
  \bibinfo{person}{Fran{\c{c}}ois Goasdou{\'{e}}}, \bibinfo{person}{Ioana
  Manolescu}, {and} \bibinfo{person}{Alexandra Roatis}.}
  \bibinfo{year}{2015}\natexlab{}.
\newblock \showarticletitle{Reasoning on web data: Algorithms and performance}.
  In \bibinfo{booktitle}{\emph{Proceedings of the {IEEE} International
  Conference on Data Engineering ({ICDE})}}. \bibinfo{pages}{1541--1544}.
\newblock
\urldef\tempurl%
\url{https://doi.org/10.1109/ICDE.2015.7113422}
\showURL{%
\tempurl}


\bibitem[\protect\citeauthoryear{Cappuzzo, Papotti, and
  Thirumuruganathan}{Cappuzzo et~al\mbox{.}}{2020}]%
        {embeddings_of_rel_datasets}
\bibfield{author}{\bibinfo{person}{Riccardo Cappuzzo}, \bibinfo{person}{Paolo
  Papotti}, {and} \bibinfo{person}{Saravanan Thirumuruganathan}.}
  \bibinfo{year}{2020}\natexlab{}.
\newblock \showarticletitle{Creating Embeddings of Heterogeneous Relational
  Datasets for Data Integration Tasks}. In
  \bibinfo{booktitle}{\emph{Proceedings of the 2020 ACM SIGMOD International
  Conference on Management of Data}} (Portland, OR, USA)
  \emph{(\bibinfo{series}{SIGMOD '20})}. \bibinfo{publisher}{Association for
  Computing Machinery}, \bibinfo{pages}{1335--1349}.
\newblock
\showISBNx{9781450367356}
\urldef\tempurl%
\url{https://doi.org/10.1145/3318464.3389742}
\showURL{%
\tempurl}


\bibitem[\protect\citeauthoryear{Carroll, Bizer, Hayes, and Stickler}{Carroll
  et~al\mbox{.}}{2005}]%
        {NamedGraphs2005}
\bibfield{author}{\bibinfo{person}{Jeremy Carroll}, \bibinfo{person}{Christian
  Bizer}, \bibinfo{person}{Patrick Hayes}, {and} \bibinfo{person}{Patrick
  Stickler}.} \bibinfo{year}{2005}\natexlab{}.
\newblock \showarticletitle{Named graphs, provenance and trust}. In
  \bibinfo{booktitle}{\emph{Proceedings of the international conference on
  World Wide Web ({WWW})}}. \bibinfo{pages}{613--622}.
\newblock
\urldef\tempurl%
\url{https://doi.org/10.1145/1060745.1060835}
\showURL{%
\tempurl}


\bibitem[\protect\citeauthoryear{Deng, Sun, Lees, Wu, and Yu}{Deng
  et~al\mbox{.}}{2020}]%
        {turl}
\bibfield{author}{\bibinfo{person}{Xiang Deng}, \bibinfo{person}{Huan Sun},
  \bibinfo{person}{Alyssa Lees}, \bibinfo{person}{You Wu}, {and}
  \bibinfo{person}{Cong Yu}.} \bibinfo{year}{2020}\natexlab{}.
\newblock \showarticletitle{TURL: Table Understanding through Representation
  Learning}.
\newblock \bibinfo{journal}{\emph{Proc. VLDB Endow.}} \bibinfo{volume}{14},
  \bibinfo{number}{3} (\bibinfo{date}{nov} \bibinfo{year}{2020}),
  \bibinfo{pages}{307--319}.
\newblock
\showISSN{2150-8097}
\urldef\tempurl%
\url{https://doi.org/10.14778/3430915.3430921}
\showDOI{\tempurl}


\bibitem[\protect\citeauthoryear{Dua and Graff}{Dua and Graff}{2017}]%
        {uci}
\bibfield{author}{\bibinfo{person}{Dheeru Dua} {and} \bibinfo{person}{Casey
  Graff}.} \bibinfo{year}{2017}\natexlab{}.
\newblock \bibinfo{title}{{UCI} Machine Learning Repository}.
\newblock
\newblock
\urldef\tempurl%
\url{http://archive.ics.uci.edu/ml}
\showURL{%
\tempurl}


\bibitem[\protect\citeauthoryear{Fan, Wang, Li, Zhang, and Miller}{Fan
  et~al\mbox{.}}{2023}]%
        {starmie}
\bibfield{author}{\bibinfo{person}{Grace Fan}, \bibinfo{person}{Jin Wang},
  \bibinfo{person}{Yuliang Li}, \bibinfo{person}{Dan Zhang}, {and}
  \bibinfo{person}{Ren{\'{e}}e Miller}.} \bibinfo{year}{2023}\natexlab{}.
\newblock \showarticletitle{Semantics-Aware Dataset Discovery from Data Lakes
  with Contextualized Column-Based Representation Learning}.
\newblock \bibinfo{journal}{\emph{Proc. VLDB Endow.}} \bibinfo{volume}{16},
  \bibinfo{number}{7} (\bibinfo{year}{2023}), \bibinfo{pages}{1726--1739}.
\newblock
\showISSN{2150-8097}
\urldef\tempurl%
\url{https://doi.org/10.14778/3587136.3587146}
\showURL{%
\tempurl}


\bibitem[\protect\citeauthoryear{Fernandez, Abedjan, Koko, Yuan, Madden, and
  et~al.}{Fernandez et~al\mbox{.}}{2018}]%
        {aurum}
\bibfield{author}{\bibinfo{person}{Raul Fernandez}, \bibinfo{person}{Ziawasch
  Abedjan}, \bibinfo{person}{Famien Koko}, \bibinfo{person}{Gina Yuan},
  \bibinfo{person}{Samuel Madden}, {and} \bibinfo{person}{et al.}}
  \bibinfo{year}{2018}\natexlab{}.
\newblock \showarticletitle{Aurum: {A} Data Discovery System}. In
  \bibinfo{booktitle}{\emph{Proceedings of International Conference on Data
  Engineering ({ICDE})}}. \bibinfo{pages}{1001--1012}.
\newblock
\urldef\tempurl%
\url{https://doi.org/10.1109/ICDE.2018.00094}
\showURL{%
\tempurl}


\bibitem[\protect\citeauthoryear{Feurer, Klein, Eggensperger, Springenberg,
  Blum, and Hutter}{Feurer et~al\mbox{.}}{2015}]%
        {autosklearn}
\bibfield{author}{\bibinfo{person}{Matthias Feurer}, \bibinfo{person}{Aaron
  Klein}, \bibinfo{person}{Katharina Eggensperger}, \bibinfo{person}{Jost
  Springenberg}, \bibinfo{person}{Manuel Blum}, {and} \bibinfo{person}{Frank
  Hutter}.} \bibinfo{year}{2015}\natexlab{}.
\newblock \showarticletitle{Efficient and Robust Automated Machine Learning}.
  In \bibinfo{booktitle}{\emph{Proceedings of the International Conference on
  Neural Information Processing Systems (NeurIPS)}}.
  \bibinfo{pages}{2962--2970}.
\newblock
\urldef\tempurl%
\url{https://dl.acm.org/doi/10.5555/2969442.2969547}
\showURL{%
\tempurl}


\bibitem[\protect\citeauthoryear{Gal{\'{a}}rraga, Teflioudi, Hose, and
  Suchanek}{Gal{\'{a}}rraga et~al\mbox{.}}{2013}]%
        {AMIE13}
\bibfield{author}{\bibinfo{person}{Luis Gal{\'{a}}rraga},
  \bibinfo{person}{Christina Teflioudi}, \bibinfo{person}{Katja Hose}, {and}
  \bibinfo{person}{Fabian Suchanek}.} \bibinfo{year}{2013}\natexlab{}.
\newblock \showarticletitle{{AMIE:} association rule mining under incomplete
  evidence in ontological knowledge bases}. In
  \bibinfo{booktitle}{\emph{Proceedings of the International World Wide Web
  Conference ({WWW})}}. \bibinfo{pages}{413--422}.
\newblock
\urldef\tempurl%
\url{https://doi.org/10.1145/2488388.2488425}
\showURL{%
\tempurl}


\bibitem[\protect\citeauthoryear{Gal{\'{a}}rraga, Teflioudi, Hose, and
  Suchanek}{Gal{\'{a}}rraga et~al\mbox{.}}{2015}]%
        {GalarragaTHS15}
\bibfield{author}{\bibinfo{person}{Luis Gal{\'{a}}rraga},
  \bibinfo{person}{Christina Teflioudi}, \bibinfo{person}{Katja Hose}, {and}
  \bibinfo{person}{Fabian~M. Suchanek}.} \bibinfo{year}{2015}\natexlab{}.
\newblock \showarticletitle{{Fast rule mining in ontological knowledge bases
  with {AMIE+}}}.
\newblock \bibinfo{journal}{\emph{Proceedings of the VLDB J. Endowment,
  (VLDB)}} (\bibinfo{year}{2015}), \bibinfo{pages}{707--730}.
\newblock
\urldef\tempurl%
\url{https://doi.org/10.1007/s00778-015-0394-1}
\showURL{%
\tempurl}


\bibitem[\protect\citeauthoryear{Gao, Zheng, Li, Li, Qin, Piao, Quan, Chang,
  Jin, He, and Li}{Gao et~al\mbox{.}}{2023}]%
        {gao2023survey}
\bibfield{author}{\bibinfo{person}{Chen Gao}, \bibinfo{person}{Yu Zheng},
  \bibinfo{person}{Nian Li}, \bibinfo{person}{Yinfeng Li},
  \bibinfo{person}{Yingrong Qin}, \bibinfo{person}{Jinghua Piao},
  \bibinfo{person}{Yuhan Quan}, \bibinfo{person}{Jianxin Chang},
  \bibinfo{person}{Depeng Jin}, \bibinfo{person}{Xiangnan He}, {and}
  \bibinfo{person}{Yong Li}.} \bibinfo{year}{2023}\natexlab{}.
\newblock \showarticletitle{A Survey of Graph Neural Networks for Recommender
  Systems: Challenges, Methods, and Directions}.
\newblock \bibinfo{journal}{\emph{ACM Trans. Recomm. Syst.}}
  \bibinfo{volume}{1}, \bibinfo{number}{1}, Article \bibinfo{articleno}{3}
  (\bibinfo{date}{mar} \bibinfo{year}{2023}), \bibinfo{numpages}{51}~pages.
\newblock
\urldef\tempurl%
\url{https://doi.org/10.1145/3568022}
\showDOI{\tempurl}


\bibitem[\protect\citeauthoryear{Goikoetxea, Agirre, and Soroa}{Goikoetxea
  et~al\mbox{.}}{2016}]%
        {semantic_model}
\bibfield{author}{\bibinfo{person}{Josu Goikoetxea}, \bibinfo{person}{Eneko
  Agirre}, {and} \bibinfo{person}{Aitor Soroa}.}
  \bibinfo{year}{2016}\natexlab{}.
\newblock \showarticletitle{Single or Multiple? Combining Word Representations
  Independently Learned from Text and WordNet}. In
  \bibinfo{booktitle}{\emph{Proceedings of the Thirtieth Conference on
  Artificial Intelligence ({AAAI})}}. \bibinfo{pages}{2608--2614}.
\newblock
\urldef\tempurl%
\url{http://www.aaai.org/ocs/index.php/AAAI/AAAI16/paper/view/11777}
\showURL{%
\tempurl}


\bibitem[\protect\citeauthoryear{Good}{Good}{1952}]%
        {cross_entropy}
\bibfield{author}{\bibinfo{person}{I.~J. Good}.}
  \bibinfo{year}{1952}\natexlab{}.
\newblock \showarticletitle{Rational Decisions}.
\newblock \bibinfo{journal}{\emph{Journal of the Royal Statistical Society.
  Series B (Methodological)}} \bibinfo{volume}{14}, \bibinfo{number}{1}
  (\bibinfo{year}{1952}), \bibinfo{pages}{107--114}.
\newblock
\showISSN{00359246}
\urldef\tempurl%
\url{http://www.jstor.org/stable/2984087}
\showURL{%
\tempurl}


\bibitem[\protect\citeauthoryear{Hartig}{Hartig}{2019}]%
        {rdfstar}
\bibfield{author}{\bibinfo{person}{Olaf Hartig}.}
  \bibinfo{year}{2019}\natexlab{}.
\newblock \showarticletitle{Foundations to Query Labeled Property Graphs using
  {SPARQL}}. In \bibinfo{booktitle}{\emph{Joint Proceedings of the 1st
  International Workshop On Semantics For Transport and the 1st International
  Workshop on Approaches for Making Data Interoperable co-located with 15th
  Semantics Conference (SEMANTiCS)}}.
\newblock
\urldef\tempurl%
\url{http://ceur-ws.org/Vol-2447/paper3.pdf}
\showURL{%
\tempurl}


\bibitem[\protect\citeauthoryear{Helal, Helali, Ammar, and Mansour}{Helal
  et~al\mbox{.}}{2021}]%
        {KGLacDemo}
\bibfield{author}{\bibinfo{person}{Ahmed Helal}, \bibinfo{person}{Mossad
  Helali}, \bibinfo{person}{Khaled Ammar}, {and} \bibinfo{person}{Essam
  Mansour}.} \bibinfo{year}{2021}\natexlab{}.
\newblock \showarticletitle{A Demonstration of {KGLac}: A Data Discovery and
  Enrichment Platform for Data Science}.
\newblock \bibinfo{journal}{\emph{Proceedings of VLDB Endowment, (PVLDB)}}
  \bibinfo{number}{12} (\bibinfo{year}{2021}), \bibinfo{pages}{2075--2089}.
\newblock
\urldef\tempurl%
\url{http://www.vldb.org/pvldb/vol13/p2075-christodoulakis.pdf}
\showURL{%
\tempurl}


\bibitem[\protect\citeauthoryear{Helali, Mansour, Abdelaziz, Dolby, and
  Srinivas}{Helali et~al\mbox{.}}{2022}]%
        {kgpip}
\bibfield{author}{\bibinfo{person}{Mossad Helali}, \bibinfo{person}{Essam
  Mansour}, \bibinfo{person}{Ibrahim Abdelaziz}, \bibinfo{person}{Julian
  Dolby}, {and} \bibinfo{person}{Kavitha Srinivas}.}
  \bibinfo{year}{2022}\natexlab{}.
\newblock \showarticletitle{A Scalable AutoML Approach Based on Graph Neural
  Networks}.
\newblock \bibinfo{journal}{\emph{Proceedings of the VLDB Endowment}}
  \bibinfo{volume}{15}, \bibinfo{number}{11} (\bibinfo{year}{2022}),
  \bibinfo{pages}{2428--2436}.
\newblock
\urldef\tempurl%
\url{https://doi.org/10.14778/3551793.3551804}
\showDOI{\tempurl}


\bibitem[\protect\citeauthoryear{Huang, Khetan, Cvitkovic, and Karnin}{Huang
  et~al\mbox{.}}{2020}]%
        {tabtransformer}
\bibfield{author}{\bibinfo{person}{Xin Huang}, \bibinfo{person}{Ashish Khetan},
  \bibinfo{person}{Milan Cvitkovic}, {and} \bibinfo{person}{Zohar Karnin}.}
  \bibinfo{year}{2020}\natexlab{}.
\newblock \bibinfo{title}{TabTransformer: Tabular Data Modeling Using
  Contextual Embeddings}.
\newblock
\newblock
\showeprint[arxiv]{2012.06678}


\bibitem[\protect\citeauthoryear{Hulsebos, Hu, Bakker, Zgraggen, Satyanarayan,
  Kraska, Demiralp, and Hidalgo}{Hulsebos et~al\mbox{.}}{2019}]%
        {sherlock}
\bibfield{author}{\bibinfo{person}{Madelon Hulsebos}, \bibinfo{person}{Kevin
  Hu}, \bibinfo{person}{Michiel Bakker}, \bibinfo{person}{Emanuel Zgraggen},
  \bibinfo{person}{Arvind Satyanarayan}, \bibinfo{person}{Tim Kraska},
  \bibinfo{person}{{\c{C}a\u{g}atay} Demiralp}, {and}
  \bibinfo{person}{{C\'{e}sar} Hidalgo}.} \bibinfo{year}{2019}\natexlab{}.
\newblock \showarticletitle{{Sherlock: A Deep Learning Approach to Semantic
  Data Type Detection}}. In \bibinfo{booktitle}{\emph{ACM Knowledge Discovery
  and Data Mining (KDD)}}.
\newblock
\urldef\tempurl%
\url{https://doi.org/10.1145/3292500.3330993}
\showDOI{\tempurl}


\bibitem[\protect\citeauthoryear{Iida, Thai, Manjunatha, and Iyyer}{Iida
  et~al\mbox{.}}{2021}]%
        {tabbie}
\bibfield{author}{\bibinfo{person}{Hiroshi Iida}, \bibinfo{person}{Dung Thai},
  \bibinfo{person}{Varun Manjunatha}, {and} \bibinfo{person}{Mohit Iyyer}.}
  \bibinfo{year}{2021}\natexlab{}.
\newblock \showarticletitle{{TABBIE}: Pretrained Representations of Tabular
  Data}. In \bibinfo{booktitle}{\emph{Proceedings of the 2021 Conference of the
  North American Chapter of the Association for Computational Linguistics:
  Human Language Technologies}}, \bibfield{editor}{\bibinfo{person}{Kristina
  Toutanova}, \bibinfo{person}{Anna Rumshisky}, \bibinfo{person}{Luke
  Zettlemoyer}, \bibinfo{person}{Dilek Hakkani-Tur},
  \bibinfo{person}{Iz~Beltagy}, \bibinfo{person}{Steven Bethard},
  \bibinfo{person}{Ryan Cotterell}, \bibinfo{person}{Tanmoy Chakraborty}, {and}
  \bibinfo{person}{Yichao Zhou}} (Eds.). \bibinfo{publisher}{Association for
  Computational Linguistics}, \bibinfo{pages}{3446--3456}.
\newblock
\urldef\tempurl%
\url{https://doi.org/10.18653/v1/2021.naacl-main.270}
\showURL{%
\tempurl}


\bibitem[\protect\citeauthoryear{Jiomekong, Camara, and Tchuente}{Jiomekong
  et~al\mbox{.}}{2019}]%
        {hmm_java_ontology}
\bibfield{author}{\bibinfo{person}{Azanzi Jiomekong}, \bibinfo{person}{Gaoussou
  Camara}, {and} \bibinfo{person}{Maurice Tchuente}.}
  \bibinfo{year}{2019}\natexlab{}.
\newblock \showarticletitle{Extracting ontological knowledge from Java source
  code using Hidden Markov Models}.
\newblock \bibinfo{journal}{\emph{Open Computer Science}}
  (\bibinfo{year}{2019}), \bibinfo{pages}{181--199}.
\newblock
\urldef\tempurl%
\url{https://doi.org/10.1515/comp-2019-0013}
\showURL{%
\tempurl}


\bibitem[\protect\citeauthoryear{Johnson, Douze, and J{\'e}gou}{Johnson
  et~al\mbox{.}}{2021}]%
        {faiss}
\bibfield{author}{\bibinfo{person}{Jeff Johnson}, \bibinfo{person}{Matthijs
  Douze}, {and} \bibinfo{person}{Herv{\'e} J{\'e}gou}.}
  \bibinfo{year}{2021}\natexlab{}.
\newblock \showarticletitle{Billion-scale similarity search with {GPUs}}.
\newblock \bibinfo{journal}{\emph{Proceedings of IEEE Transactions on Big
  Data}} \bibinfo{number}{3} (\bibinfo{year}{2021}), \bibinfo{pages}{535--547}.
\newblock
\urldef\tempurl%
\url{https://doi.org/10.1109/TBDATA.2019.2921572}
\showURL{%
\tempurl}


\bibitem[\protect\citeauthoryear{Kaul, Maheshwary, and Pudi}{Kaul
  et~al\mbox{.}}{2017}]%
        {autolearn}
\bibfield{author}{\bibinfo{person}{Ambika Kaul}, \bibinfo{person}{Saket
  Maheshwary}, {and} \bibinfo{person}{Vikram Pudi}.}
  \bibinfo{year}{2017}\natexlab{}.
\newblock \showarticletitle{AutoLearn - Automated Feature Generation and
  Selection}. In \bibinfo{booktitle}{\emph{2017 {IEEE} International Conference
  on Data Mining, {ICDM} 2017, New Orleans, LA, USA, November 18-21, 2017}},
  \bibfield{editor}{\bibinfo{person}{Vijay Raghavan}, \bibinfo{person}{Srinivas
  Aluru}, \bibinfo{person}{George Karypis}, \bibinfo{person}{Lucio Miele},
  {and} \bibinfo{person}{Xindong Wu}} (Eds.). \bibinfo{publisher}{{IEEE}
  Computer Society}, \bibinfo{pages}{217--226}.
\newblock
\urldef\tempurl%
\url{https://doi.org/10.1109/ICDM.2017.31}
\showDOI{\tempurl}


\bibitem[\protect\citeauthoryear{Khatiwada, Fan, Shraga, Chen, Gatterbauer,
  Miller, and Riedewald}{Khatiwada et~al\mbox{.}}{2023}]%
        {santos}
\bibfield{author}{\bibinfo{person}{Aamod Khatiwada}, \bibinfo{person}{Grace
  Fan}, \bibinfo{person}{Roee Shraga}, \bibinfo{person}{Zixuan Chen},
  \bibinfo{person}{Wolfgang Gatterbauer}, \bibinfo{person}{Ren{\'e}e~J Miller},
  {and} \bibinfo{person}{Mirek Riedewald}.} \bibinfo{year}{2023}\natexlab{}.
\newblock \showarticletitle{SANTOS: Relationship-based Semantic Table Union
  Search}. In \bibinfo{booktitle}{\emph{SIGMOD Conference 2023}}.
  \bibinfo{publisher}{ACM}.
\newblock
\urldef\tempurl%
\url{https://doi.org/10.1145/3588689}
\showURL{%
\tempurl}


\bibitem[\protect\citeauthoryear{Klyne, Carrol, and McBride}{Klyne
  et~al\mbox{.}}{2014}]%
        {klyne2014rdf}
\bibfield{author}{\bibinfo{person}{Graham Klyne}, \bibinfo{person}{Jeremy
  Carrol}, {and} \bibinfo{person}{Brian McBride}.}
  \bibinfo{year}{2014}\natexlab{}.
\newblock \bibinfo{title}{{RDF 1.1 Concepts and Abstract Syntax. World-Wide Web
  Consortium}}.
\newblock
\newblock
\urldef\tempurl%
\url{https://www.w3.org/TR/rdf11-concepts/}
\showURL{%
\tempurl}


\bibitem[\protect\citeauthoryear{Malkov and Yashunin}{Malkov and
  Yashunin}{2020}]%
        {hnsw}
\bibfield{author}{\bibinfo{person}{Yu~A. Malkov} {and} \bibinfo{person}{D.~A.
  Yashunin}.} \bibinfo{year}{2020}\natexlab{}.
\newblock \showarticletitle{Efficient and Robust Approximate Nearest Neighbor
  Search Using Hierarchical Navigable Small World Graphs}.
\newblock \bibinfo{journal}{\emph{IEEE Transactions on Pattern Analysis and
  Machine Intelligence}} \bibinfo{volume}{42}, \bibinfo{number}{4}
  (\bibinfo{year}{2020}), \bibinfo{pages}{824--836}.
\newblock
\urldef\tempurl%
\url{https://doi.org/10.1109/TPAMI.2018.2889473}
\showURL{%
\tempurl}


\bibitem[\protect\citeauthoryear{Mansour, Srinivas, and Hose}{Mansour
  et~al\mbox{.}}{2021}]%
        {KEK}
\bibfield{author}{\bibinfo{person}{Essam Mansour}, \bibinfo{person}{Kavitha
  Srinivas}, {and} \bibinfo{person}{Katja Hose}.}
  \bibinfo{year}{2021}\natexlab{}.
\newblock \showarticletitle{Federated Data Science to Break Down Silos
  [Vision]}.
\newblock \bibinfo{journal}{\emph{{SIGMOD} Record}} \bibinfo{volume}{50},
  \bibinfo{number}{4} (\bibinfo{year}{2021}), \bibinfo{pages}{16--22}.
\newblock
\urldef\tempurl%
\url{https://doi.org/10.1145/3516431.3516435}
\showDOI{\tempurl}


\bibitem[\protect\citeauthoryear{Montoya, Skaf{-}Molli, and Hose}{Montoya
  et~al\mbox{.}}{2017}]%
        {MontoyaSH17}
\bibfield{author}{\bibinfo{person}{Gabriela Montoya}, \bibinfo{person}{Hala
  Skaf{-}Molli}, {and} \bibinfo{person}{Katja Hose}.}
  \bibinfo{year}{2017}\natexlab{}.
\newblock \showarticletitle{{The Odyssey Approach for Optimizing Federated
  {SPARQL} Queries}}. In \bibinfo{booktitle}{\emph{Proceedings of the
  International Semantic Web Conference ({ISWC})}},
  Vol.~\bibinfo{volume}{10587}. \bibinfo{pages}{471--489}.
\newblock
\urldef\tempurl%
\url{https://doi.org/10.1007/978-3-319-68288-4\_28}
\showURL{%
\tempurl}


\bibitem[\protect\citeauthoryear{Mueller and Smola}{Mueller and Smola}{2019}]%
        {MuellerS19}
\bibfield{author}{\bibinfo{person}{Jonas Mueller} {and} \bibinfo{person}{Alex
  Smola}.} \bibinfo{year}{2019}\natexlab{}.
\newblock \showarticletitle{Recognizing Variables from Their Data via Deep
  Embeddings of Distributions}. In \bibinfo{booktitle}{\emph{International
  Conference on Data Mining ({ICDM})}}. \bibinfo{pages}{1264--1269}.
\newblock
\urldef\tempurl%
\url{https://doi.org/10.1109/ICDM.2019.00158}
\showURL{%
\tempurl}


\bibitem[\protect\citeauthoryear{Nargesian, Samulowitz, Khurana, Khalil, and
  Turaga}{Nargesian et~al\mbox{.}}{2017}]%
        {lfe}
\bibfield{author}{\bibinfo{person}{Fatemeh Nargesian}, \bibinfo{person}{Horst
  Samulowitz}, \bibinfo{person}{Udayan Khurana}, \bibinfo{person}{Elias~B.
  Khalil}, {and} \bibinfo{person}{Deepak~S. Turaga}.}
  \bibinfo{year}{2017}\natexlab{}.
\newblock \showarticletitle{Learning Feature Engineering for Classification}.
  In \bibinfo{booktitle}{\emph{Proceedings of the Twenty-Sixth International
  Joint Conference on Artificial Intelligence}},
  \bibfield{editor}{\bibinfo{person}{Carles Sierra}} (Ed.).
  \bibinfo{pages}{2529--2535}.
\newblock
\urldef\tempurl%
\url{https://doi.org/10.24963/ijcai.2017/352}
\showURL{%
\tempurl}


\bibitem[\protect\citeauthoryear{Nargesian, Zhu, Pu, and Miller}{Nargesian
  et~al\mbox{.}}{2018}]%
        {Fatemeh18}
\bibfield{author}{\bibinfo{person}{Fatemeh Nargesian}, \bibinfo{person}{Erkang
  Zhu}, \bibinfo{person}{Ken Pu}, {and} \bibinfo{person}{Ren{\'{e}}e Miller}.}
  \bibinfo{year}{2018}\natexlab{}.
\newblock \showarticletitle{Table Union Search on Open Data}.
\newblock \bibinfo{journal}{\emph{Proceedings of the {VLDB} Endowment,
  (PVLDB)}} (\bibinfo{year}{2018}), \bibinfo{pages}{813--825}.
\newblock
\urldef\tempurl%
\url{http://www.vldb.org/pvldb/vol11/p813-nargesian.pdf}
\showURL{%
\tempurl}


\bibitem[\protect\citeauthoryear{Peng, Wu, Lockhart, and et~al}{Peng
  et~al\mbox{.}}{2021}]%
        {dataprep}
\bibfield{author}{\bibinfo{person}{Jinglin Peng}, \bibinfo{person}{Weiyuan Wu},
  \bibinfo{person}{Brandon Lockhart}, {and} \bibinfo{person}{et al}.}
  \bibinfo{year}{2021}\natexlab{}.
\newblock \showarticletitle{DataPrep.EDA: Task-Centric Exploratory Data
  Analysis for Statistical Modeling in Python}. In
  \bibinfo{booktitle}{\emph{SIGMOD}}. \bibinfo{pages}{2271--2280}.
\newblock
\urldef\tempurl%
\url{https://doi.org/10.1145/3448016.3457330}
\showURL{%
\tempurl}


\bibitem[\protect\citeauthoryear{Pennington, Socher, and Manning}{Pennington
  et~al\mbox{.}}{2014}]%
        {glove}
\bibfield{author}{\bibinfo{person}{Jeffrey Pennington},
  \bibinfo{person}{Richard Socher}, {and} \bibinfo{person}{Christopher
  Manning}.} \bibinfo{year}{2014}\natexlab{}.
\newblock \showarticletitle{{G}lo{V}e: Global Vectors for Word Representation}.
  In \bibinfo{booktitle}{\emph{Proceedings of the 2014 Conference on Empirical
  Methods in Natural Language Processing ({EMNLP})}}.
  \bibinfo{pages}{1532--1543}.
\newblock
\urldef\tempurl%
\url{https://doi.org/10.3115/v1/D14-1162}
\showURL{%
\tempurl}


\bibitem[\protect\citeauthoryear{Peters, Ammar, Bhagavatula, and Power}{Peters
  et~al\mbox{.}}{2017}]%
        {ner_model}
\bibfield{author}{\bibinfo{person}{Matthew Peters}, \bibinfo{person}{Waleed
  Ammar}, \bibinfo{person}{Chandra Bhagavatula}, {and} \bibinfo{person}{Russell
  Power}.} \bibinfo{year}{2017}\natexlab{}.
\newblock \showarticletitle{Semi-supervised sequence tagging with bidirectional
  language models}. In \bibinfo{booktitle}{\emph{Proceedings of the Association
  for Computational Linguistics (ACL)}}, Vol.~\bibinfo{volume}{1}.
  \bibinfo{pages}{1756--1765}.
\newblock
\urldef\tempurl%
\url{https://doi.org/10.18653/v1/P17-1161}
\showURL{%
\tempurl}


\bibitem[\protect\citeauthoryear{Petersohn}{Petersohn}{2021}]%
        {Petersohn21}
\bibfield{author}{\bibinfo{person}{Devin Petersohn}.}
  \bibinfo{year}{2021}\natexlab{}.
\newblock \showarticletitle{Scaling Data Science does not mean Scaling
  Machines}. In \bibinfo{booktitle}{\emph{Conference on Innovative Data Systems
  Research ({CIDR})}}.
\newblock
\urldef\tempurl%
\url{http://cidrdb.org/cidr2021/papers/cidr2021\_abstract11.pdf}
\showURL{%
\tempurl}


\bibitem[\protect\citeauthoryear{Raju, Lakshmi, Jain, Kalidindi, and
  Padma}{Raju et~al\mbox{.}}{2020}]%
        {scaling_helps}
\bibfield{author}{\bibinfo{person}{V~N~Ganapathi Raju},
  \bibinfo{person}{K~Prasanna Lakshmi}, \bibinfo{person}{Vinod~Mahesh Jain},
  \bibinfo{person}{Archana Kalidindi}, {and} \bibinfo{person}{V Padma}.}
  \bibinfo{year}{2020}\natexlab{}.
\newblock \showarticletitle{Study the Influence of Normalization/Transformation
  process on the Accuracy of Supervised Classification}. In
  \bibinfo{booktitle}{\emph{2020 Third International Conference on Smart
  Systems and Inventive Technology (ICSSIT)}}. \bibinfo{pages}{729--735}.
\newblock
\urldef\tempurl%
\url{https://doi.org/10.1109/ICSSIT48917.2020.9214160}
\showDOI{\tempurl}


\bibitem[\protect\citeauthoryear{Rekatsinas, Chu, Ilyas, and R{\'e}}{Rekatsinas
  et~al\mbox{.}}{2017}]%
        {HoloClean}
\bibfield{author}{\bibinfo{person}{Theodoros Rekatsinas}, \bibinfo{person}{Xu
  Chu}, \bibinfo{person}{Ihab~F. Ilyas}, {and} \bibinfo{person}{Christopher
  R{\'e}}.} \bibinfo{year}{2017}\natexlab{}.
\newblock \showarticletitle{HoloClean: Holistic Data Repairs with Probabilistic
  Inference}.
\newblock \bibinfo{journal}{\emph{Proc. VLDB Endow.}} \bibinfo{volume}{10},
  \bibinfo{number}{11} (\bibinfo{year}{2017}), \bibinfo{pages}{1190--1201}.
\newblock
\showISSN{2150-8097}
\urldef\tempurl%
\url{https://doi.org/10.14778/3137628.3137631}
\showURL{%
\tempurl}


\bibitem[\protect\citeauthoryear{Rezig, Ouzzani, Aref, and et~al.}{Rezig
  et~al\mbox{.}}{2021}]%
        {kindi2021}
\bibfield{author}{\bibinfo{person}{El~Kindi Rezig}, \bibinfo{person}{Mourad
  Ouzzani}, \bibinfo{person}{Walid~G. Aref}, {and} \bibinfo{person}{et al.}}
  \bibinfo{year}{2021}\natexlab{}.
\newblock \showarticletitle{Horizon: Scalable Dependency-driven Data Cleaning}.
\newblock \bibinfo{journal}{\emph{{PVLDB}}} \bibinfo{volume}{14},
  \bibinfo{number}{11} (\bibinfo{year}{2021}).
\newblock


\bibitem[\protect\citeauthoryear{Salis, Sotiropoulos, Louridas, Spinellis, and
  Mitropoulos}{Salis et~al\mbox{.}}{2021}]%
        {pycg}
\bibfield{author}{\bibinfo{person}{Vitalis Salis}, \bibinfo{person}{Thodoris
  Sotiropoulos}, \bibinfo{person}{Panos Louridas}, \bibinfo{person}{Diomidis
  Spinellis}, {and} \bibinfo{person}{Dimitris Mitropoulos}.}
  \bibinfo{year}{2021}\natexlab{}.
\newblock \showarticletitle{PyCG: Practical Call Graph Generation in Python}.
\newblock \bibinfo{journal}{\emph{International Conference on Software
  Engineering ({ICSE})}} (\bibinfo{year}{2021}), \bibinfo{pages}{1646--1657}.
\newblock
\urldef\tempurl%
\url{https://doi.org/10.1109/ICSE43902.2021.00146}
\showURL{%
\tempurl}


\bibitem[\protect\citeauthoryear{Schwarte, Haase, Hose, Schenkel, and
  Schmidt}{Schwarte et~al\mbox{.}}{2011}]%
        {SchwarteHHSS11}
\bibfield{author}{\bibinfo{person}{Andreas Schwarte}, \bibinfo{person}{Peter
  Haase}, \bibinfo{person}{Katja Hose}, \bibinfo{person}{Ralf Schenkel}, {and}
  \bibinfo{person}{Michael Schmidt}.} \bibinfo{year}{2011}\natexlab{}.
\newblock \showarticletitle{FedX: {A} Federation Layer for Distributed Query
  Processing on Linked Open Data}. In \bibinfo{booktitle}{\emph{Proceedings of
  The Semanic Web: Research and Applications (ESWC)}}
  \emph{(\bibinfo{series}{Lecture Notes in Computer Science})}.
  \bibinfo{publisher}{Springer}, \bibinfo{pages}{481--486}.
\newblock
\urldef\tempurl%
\url{https://doi.org/10.1007/978-3-642-21064-8\_39}
\showURL{%
\tempurl}


\bibitem[\protect\citeauthoryear{Suchanek, Kasneci, and Weikum}{Suchanek
  et~al\mbox{.}}{2007}]%
        {yago}
\bibfield{author}{\bibinfo{person}{Fabian~M. Suchanek},
  \bibinfo{person}{Gjergji Kasneci}, {and} \bibinfo{person}{Gerhard Weikum}.}
  \bibinfo{year}{2007}\natexlab{}.
\newblock \showarticletitle{Yago: A Core of Semantic Knowledge}. In
  \bibinfo{booktitle}{\emph{Proceedings of the 16th International Conference on
  World Wide Web}}. \bibinfo{pages}{697--706}.
\newblock
\showISBNx{9781595936547}
\urldef\tempurl%
\url{https://doi.org/10.1145/1242572.1242667}
\showURL{%
\tempurl}


\bibitem[\protect\citeauthoryear{Vanschoren, Rijn, Bischl, and
  Torgo}{Vanschoren et~al\mbox{.}}{2013}]%
        {OpenML}
\bibfield{author}{\bibinfo{person}{Joaquin Vanschoren}, \bibinfo{person}{Jan
  Rijn}, \bibinfo{person}{Bernd Bischl}, {and} \bibinfo{person}{Luis Torgo}.}
  \bibinfo{year}{2013}\natexlab{}.
\newblock \showarticletitle{{OpenML}: Networked Science in Machine Learning}.
\newblock \bibinfo{journal}{\emph{SIGKDD Explorations}}  \bibinfo{volume}{15}
  (\bibinfo{year}{2013}), \bibinfo{pages}{49--60}.
\newblock
\urldef\tempurl%
\url{https://doi.org/10.1145/2641190.2641198}
\showURL{%
\tempurl}


\bibitem[\protect\citeauthoryear{Villaz{\'{o}}n{-}Terrazas,
  Garc{\'{\i}}a{-}Santa, Ren, Srinivas, Rodriguez{-}Muro, and
  et~al.}{Villaz{\'{o}}n{-}Terrazas et~al\mbox{.}}{2017}]%
        {Villazon-TerrazasGRSRAP17}
\bibfield{author}{\bibinfo{person}{Boris Villaz{\'{o}}n{-}Terrazas},
  \bibinfo{person}{Nuria Garc{\'{\i}}a{-}Santa}, \bibinfo{person}{Yuan Ren},
  \bibinfo{person}{Kavitha Srinivas}, \bibinfo{person}{Mariano
  Rodriguez{-}Muro}, {and} \bibinfo{person}{et al.}}
  \bibinfo{year}{2017}\natexlab{}.
\newblock \showarticletitle{Construction of Enterprise Knowledge Graphs {(I)}}.
\newblock In \bibinfo{booktitle}{\emph{Proceedings of Exploiting Linked Data
  and Knowledge Graphs in Large Organisations}}. \bibinfo{pages}{87--116}.
\newblock
\urldef\tempurl%
\url{https://doi.org/10.1007/978-3-319-45654-6\_4}
\showURL{%
\tempurl}


\bibitem[\protect\citeauthoryear{Wang, Wu, Weimer, and Zhu}{Wang
  et~al\mbox{.}}{2021}]%
        {flaml}
\bibfield{author}{\bibinfo{person}{Chi Wang}, \bibinfo{person}{Qingyun Wu},
  \bibinfo{person}{Markus Weimer}, {and} \bibinfo{person}{Erkang Zhu}.}
  \bibinfo{year}{2021}\natexlab{}.
\newblock \showarticletitle{FLAML: A Fast and Lightweight AutoML Library}. In
  \bibinfo{booktitle}{\emph{Proceedings of Machine Learning and Systems
  (MLSys)}}, Vol.~\bibinfo{volume}{3}. \bibinfo{pages}{434--447}.
\newblock
\urldef\tempurl%
\url{https://proceedings.mlsys.org/paper/2021/file/92cc227532d17e56e07902b254dfad10-Paper.pdf}
\showURL{%
\tempurl}


\bibitem[\protect\citeauthoryear{Wang, Guan, Ma, Bian, Che, Daundkar,
  Sehirlioglu, and Wu}{Wang et~al\mbox{.}}{2023}]%
        {topk}
\bibfield{author}{\bibinfo{person}{Mengying Wang}, \bibinfo{person}{Sheng
  Guan}, \bibinfo{person}{Hanchao Ma}, \bibinfo{person}{Yiyang Bian},
  \bibinfo{person}{Haolai Che}, \bibinfo{person}{Abhishek Daundkar},
  \bibinfo{person}{Alp Sehirlioglu}, {and} \bibinfo{person}{Yinghui Wu}.}
  \bibinfo{year}{2023}\natexlab{}.
\newblock \showarticletitle{Selecting Top-k Data Science Models by Example
  Dataset}. \bibinfo{publisher}{International Conference on Information and
  Knowledge Management (CIKM)}, \bibinfo{pages}{2686 -- 2695}.
\newblock
\urldef\tempurl%
\url{https://doi.org/10.1145/3583780.3615051}
\showURL{%
\tempurl}


\bibitem[\protect\citeauthoryear{Weischedel, Palmer, Marcus, Hovy, Pradhan,
  Ramshaw, Xue, Taylor, Kaufman, Franchini, El-Bachouti, Belvin, and
  Houston}{Weischedel et~al\mbox{.}}{2013}]%
        {ontonotes}
\bibfield{author}{\bibinfo{person}{Ralph Weischedel}, \bibinfo{person}{Martha
  Palmer}, \bibinfo{person}{Mitchell Marcus}, \bibinfo{person}{Eduard Hovy},
  \bibinfo{person}{Sameer Pradhan}, \bibinfo{person}{Lance Ramshaw},
  \bibinfo{person}{Nianwen Xue}, \bibinfo{person}{Ann Taylor},
  \bibinfo{person}{Jeff Kaufman}, \bibinfo{person}{Michelle Franchini},
  \bibinfo{person}{Mohammed El-Bachouti}, \bibinfo{person}{Robert Belvin},
  {and} \bibinfo{person}{Ann Houston}.} \bibinfo{year}{2013}\natexlab{}.
\newblock \bibinfo{title}{{OntoNotes Release 5.0 LDC2013T19}}.
\newblock \bibinfo{howpublished}{Web Download}.
\newblock
\urldef\tempurl%
\url{https://doi.org/10.35111/xmhb-2b84}
\showURL{%
\tempurl}


\bibitem[\protect\citeauthoryear{Wu, Zhang, Ilyas, and Rekatsinas}{Wu
  et~al\mbox{.}}{2020}]%
        {aimnet}
\bibfield{author}{\bibinfo{person}{Richard Wu}, \bibinfo{person}{Aoqian Zhang},
  \bibinfo{person}{Ihab~F. Ilyas}, {and} \bibinfo{person}{Theodoros
  Rekatsinas}.} \bibinfo{year}{2020}\natexlab{}.
\newblock \showarticletitle{Attention-based Learning for Missing Data
  Imputation in HoloClean}. In \bibinfo{booktitle}{\emph{Conference on Machine
  Learning and Systems}}.
\newblock
\urldef\tempurl%
\url{https://api.semanticscholar.org/CorpusID:211482719}
\showURL{%
\tempurl}


\bibitem[\protect\citeauthoryear{Yan and He}{Yan and He}{2020}]%
        {Auto-Suggest}
\bibfield{author}{\bibinfo{person}{Cong Yan} {and} \bibinfo{person}{Yeye He}.}
  \bibinfo{year}{2020}\natexlab{}.
\newblock \showarticletitle{{Auto-Suggest}: Learning-to-Recommend Data
  Preparation Steps Using Data Science Notebooks}. In
  \bibinfo{booktitle}{\emph{{SIGMOD}}}. \bibinfo{pages}{1539--1554}.
\newblock


\bibitem[\protect\citeauthoryear{Yang, He, and Chaudhuri}{Yang
  et~al\mbox{.}}{2021}]%
        {autopipeline}
\bibfield{author}{\bibinfo{person}{Junwen Yang}, \bibinfo{person}{Yeye He},
  {and} \bibinfo{person}{Surajit Chaudhuri}.} \bibinfo{year}{2021}\natexlab{}.
\newblock \showarticletitle{Auto-pipeline: synthesizing complex data pipelines
  by-target using reinforcement learning and search}.
\newblock \bibinfo{journal}{\emph{{PVLDB}}} \bibinfo{volume}{14},
  \bibinfo{number}{11} (\bibinfo{year}{2021}), \bibinfo{pages}{2563--2575}.
\newblock


\bibitem[\protect\citeauthoryear{Zeng, Zhou, Srivastava, Kannan, and
  Prasanna}{Zeng et~al\mbox{.}}{2020}]%
        {GraphSAINT}
\bibfield{author}{\bibinfo{person}{Hanqing Zeng}, \bibinfo{person}{Hongkuan
  Zhou}, \bibinfo{person}{Ajitesh Srivastava}, \bibinfo{person}{Rajgopal
  Kannan}, {and} \bibinfo{person}{Viktor~K. Prasanna}.}
  \bibinfo{year}{2020}\natexlab{}.
\newblock \showarticletitle{GraphSAINT: Graph Sampling Based Inductive Learning
  Method}. In \bibinfo{booktitle}{\emph{8th International Conference on
  Learning Representations, {ICLR} 2020, Addis Ababa, Ethiopia, April 26-30,
  2020}}.
\newblock
\urldef\tempurl%
\url{https://openreview.net/forum?id=BJe8pkHFwS}
\showURL{%
\tempurl}
\newblock
\shownote{, GitHub Code:
  \url{https://github.com/snap-stanford/ogb/blob/master/examples/nodeproppred/mag/graph_saint.py}.}


\bibitem[\protect\citeauthoryear{Zhang, Hulsebos, Suhara, Demiralp, Li, and
  Tan}{Zhang et~al\mbox{.}}{2020}]%
        {sato}
\bibfield{author}{\bibinfo{person}{Dan Zhang}, \bibinfo{person}{Madelon
  Hulsebos}, \bibinfo{person}{Yoshihiko Suhara},
  \bibinfo{person}{\c{C}a\u{g}atay Demiralp}, \bibinfo{person}{Jinfeng Li},
  {and} \bibinfo{person}{Wang-Chiew Tan}.} \bibinfo{year}{2020}\natexlab{}.
\newblock \showarticletitle{Sato: Contextual Semantic Type Detection in
  Tables}.
\newblock \bibinfo{journal}{\emph{Proc. VLDB Endow.}} \bibinfo{volume}{13},
  \bibinfo{number}{12} (\bibinfo{date}{jul} \bibinfo{year}{2020}),
  \bibinfo{pages}{1835--1848}.
\newblock
\showISSN{2150-8097}
\urldef\tempurl%
\url{https://doi.org/10.14778/3407790.3407793}
\showURL{%
\tempurl}


\bibitem[\protect\citeauthoryear{Zhang and Ives}{Zhang and Ives}{2020}]%
        {ZhangI20}
\bibfield{author}{\bibinfo{person}{Yi Zhang} {and} \bibinfo{person}{Zachary
  Ives}.} \bibinfo{year}{2020}\natexlab{}.
\newblock \showarticletitle{Finding Related Tables in Data Lakes for
  Interactive Data Science}. In \bibinfo{booktitle}{\emph{Proceedings of The
  International Conference on Management of Data, (SIGMOD)}}.
  \bibinfo{pages}{1951--1966}.
\newblock
\urldef\tempurl%
\url{https://doi.org/10.1145/3318464.3389726}
\showURL{%
\tempurl}


\bibitem[\protect\citeauthoryear{Zheng, Zou, Peng, Yan, Song, and et~al.}{Zheng
  et~al\mbox{.}}{2016}]%
        {ZhengZPYSZ16}
\bibfield{author}{\bibinfo{person}{Weiguo Zheng}, \bibinfo{person}{Lei Zou},
  \bibinfo{person}{Wei Peng}, \bibinfo{person}{Xifeng Yan},
  \bibinfo{person}{Shaoxu Song}, {and} \bibinfo{person}{et al.}}
  \bibinfo{year}{2016}\natexlab{}.
\newblock \showarticletitle{Semantic {SPARQL} Similarity Search Over {RDF}
  Knowledge Graphs}.
\newblock \bibinfo{journal}{\emph{Proceedings of the {VLDB} Endowment,
  (PVLDB)}} (\bibinfo{year}{2016}), \bibinfo{pages}{840--851}.
\newblock
\urldef\tempurl%
\url{http://www.vldb.org/pvldb/vol9/p840-zheng.pdf}
\showURL{%
\tempurl}


\bibitem[\protect\citeauthoryear{Zhu, Deng, Nargesian, and Miller}{Zhu
  et~al\mbox{.}}{2019}]%
        {ZhuDNM19}
\bibfield{author}{\bibinfo{person}{Erkang Zhu}, \bibinfo{person}{Dong Deng},
  \bibinfo{person}{Fatemeh Nargesian}, {and} \bibinfo{person}{Ren{\'{e}}e
  Miller}.} \bibinfo{year}{2019}\natexlab{}.
\newblock \showarticletitle{{JOSIE:} Overlap Set Similarity Search for Finding
  Joinable Tables in Data Lakes}. In \bibinfo{booktitle}{\emph{Proceedings of
  the International Conference on Management of Data ({SIGMOD})}}.
  \bibinfo{pages}{847--864}.
\newblock
\urldef\tempurl%
\url{https://doi.org/10.1145/3299869.3300065}
\showURL{%
\tempurl}


\end{thebibliography}

\end{document}